\documentclass[twoside,11pt]{article}

\usepackage{jmlr2e}

\usepackage{etoolbox}
\patchcmd{\maketitle}
 {\def\@makefnmark}
 {\def\@makefnmark{}\def\useless@macro}
 {}{}
\makeatother

\usepackage{amsmath}
\usepackage{multirow}
\usepackage{xcolor}
\usepackage{epstopdf}
\usepackage{caption}
\usepackage{subcaption}
\usepackage{bm}
\usepackage{pgfplots}
\usepackage{footnote}
\usepackage{stfloats}
\usepackage{placeins}
\usepackage{color,soul}
\usepackage{dsfont}
\usepackage{algorithm}
\usepackage{algpseudocode}
\usepackage{pbox}
\usepackage{enumerate}
\usepackage{etextools}
\usepackage{tabulary}
\usepackage{dsfont}
\usepackage{mathtools}

\newlength
\figureheight
\newlength
\figurewidth

\def\x{{\mathbf x}}
\def\X{{\mathbf X}}
\def\v{{\mathbf v}}
\def\d{{\mathbf d}}

\def\Y{{\mathbf Y}}

\def\Z{{\mathbf Z}}

\def\U{{\mathbf U}}
\def\Q{{\mathbf Q}}

\def\D{{\mathbf D}}

\def\B{{\mathbf B}}
\def\W{{\mathbf W}}

\def\P{{\mathcal{P}}}
\def\ReLU{{\text{ReLU}}}
\def\b{{\mathbf b}}
\def\S{{\mathcal{S}}}
\def\H{{\mathcal{H}}}

\def\PP{{\mathbf P}}
\def\RR{{\mathbf R}}
\def\SS{{\mathbf S}}
\def\V{{\mathbf V}}

\def\E{{\mathbf E}}

\def\gama{{\boldsymbol \gamma}}
\def\Gama{{\boldsymbol \Gamma}}
\def\Delt{{\boldsymbol \Delta}}

\def\O{{\boldsymbol \Omega}}
\def\lamda{{\boldsymbol \lambda}}
\def\Eps{{\bm{\mathbf{\mathcal{E}}}}}

\def\Po{{\text{P}_0}}
\def\Pone{{\text{P}_1}}
\def\Poi{{\text{P}_{0,\infty}}}
\def\Loi{{\ell_{0,\infty}}}
\def\Poie{{\text{P}_{0,\infty}^\Eps}}
\def\DCP{{\text{DCP}_\lamda}}
\def\DCPstar{{\text{DCP}_\lamda^{\hspace{0.04cm} \star}}}
\def\DLP{{\text{DLP}_\lamda}}
\def\DCPE{{\text{DCP}_\lamda^{\hspace{0.04cm} \Eps}}}
\def\DLPE{{\text{DLP}_\lamda^{\hspace{0.04cm} \Eps}}}
\def\Ltinf{{\ell_{2,\infty}}}
\def\pp{{\scriptscriptstyle{\PP}}}
\def\ss{{\scriptscriptstyle{\SS}}}

\def\ReLU{{\text{ReLU}}}

\mathtoolsset{showonlyrefs}

\DeclarePairedDelimiter{\ceil}{\lceil}{\rceil}

\firstpageno{1}

\begin{document}

\title{Convolutional Neural Networks Analyzed via \\ Convolutional Sparse Coding}

\author{\name Vardan Papyan* \thanks{The authors contributed equally to this work.} \email vardanp@campus.technion.ac.il \\
       \addr Department of Computer Science \\
       Technion -– Israel Institute of Technology \\
       Technion City, Haifa 32000, Israel
       \AND
       \name Yaniv Romano* \email yromano@tx.technion.ac.il \\
       \addr Department of Electrical Engineering \\
       Technion –- Israel Institute of Technology \\
       Technion City, Haifa 32000, Israel
       \AND
      \name Michael Elad \email elad@cs.technion.ac.il \\
      \addr Department of Computer Science \\
      Technion –- Israel Institute of Technology \\
      Technion City, Haifa 32000, Israel}

\editor{}

\maketitle

\begin{abstract}
Convolutional neural networks (CNN) have led to many state-of-the-art results spanning through various fields. However, a clear and profound theoretical understanding of the forward pass, the core algorithm of CNN, is still lacking. In parallel, within the wide field of sparse approximation, Convolutional Sparse Coding (CSC) has gained increasing attention in recent years. A theoretical study of this model was recently conducted, establishing it as a reliable and stable alternative to the commonly practiced patch-based processing. Herein, we propose a novel multi-layer model, ML-CSC, in which signals are assumed to emerge from a cascade of CSC layers. This is shown to be tightly connected to CNN, so much so that the forward pass of the CNN is in fact the thresholding pursuit serving the ML-CSC model. This connection brings a fresh view to CNN, as we are able to attribute to this architecture theoretical claims such as uniqueness of the representations throughout the network, and their stable estimation, all guaranteed under simple local sparsity conditions. Lastly, identifying the weaknesses in the above pursuit scheme, we propose an alternative to the forward pass, which is connected to deconvolutional, recurrent and residual networks, and has better theoretical guarantees.
\end{abstract}


\newpage

\ShortHeadings{Convolutional Neural Networks Analyzed via Convolutional Sparse Coding}{Papyan, Romano and Elad}

\section{Introduction}
Deep learning \citep{lecun2015deep}, and in particular CNN \citep{le1990handwritten,lecun1998gradient,krizhevsky2012imagenet}, has gained a copious amount of attention in recent years as it has led to many state-of-the-art results spanning through many fields -- including speech recognition \citep{bengio2003neural,hinton2012deep,mikolov2013efficient}, computer vision \citep{farabet2013learning,simonyan2014very,he2015deep}, signal and image processing \citep{gatys2015neural,ulyanov2016texture,johnson2016perceptual,dong2016image}, to name a few. In the context of CNN, the forward pass is a multi-layer scheme that provides an end-to-end mapping, from an input signal to some desired output. Each layer of this algorithm consists of three steps. The first convolves the input with a set of learned filters, resulting in a set of feature (or kernel) maps. These then undergo a point wise non-linear function, in a second step, often resulting in a sparse outcome \citep{glorot2011deep}. A third (and optional) down-sampling step, termed \emph{pooling}, is then applied on the result in order to reduce its dimensions. The output of this layer is then fed into another one, thus forming the multi-layer structure, often termed \emph{forward pass}.

Despite its marvelous empirical success, a clear and profound theoretical understanding of this scheme is still lacking. A few preliminary theoretical results were recently suggested. In \citep{mallat2012group,bruna2013invariant} the Scattering Transform was proposed, suggesting to replace the learned filters in the CNN with predefined Wavelet functions. Interestingly, the features obtained from this network were shown to be invariant to various transformations such as translations and rotations. Other works have studied the properties of deep and fully connected networks under the assumption of independent identically distributed random weights \citep{giryes2015deep,saxe2013exact,arora2014provable,dauphin2014identifying,choromanska2015loss}. In particular, in \citep{giryes2015deep} deep neural networks were proven to preserve the metric structure of the input data as it propagates through the layers of the network. This, in turn, was shown to allow a stable recovery of the data from the features obtained from the network.

Another prominent paradigm in data processing is the sparse representation concept, being one of the most popular choices for a prior in the signal and image processing communities, and leading to exceptional results in various applications \citep{Elad2006,Dong2011,zhang2010discriminative,jiang2011learning,Mairal2014}. In this framework, one assumes that a signal can be represented as a linear combination of a few columns (called atoms) from a matrix termed a dictionary. Put differently, the signal is equal to a multiplication of a dictionary by a sparse vector. The task of retrieving the sparsest representation of a signal over a dictionary is called sparse coding or pursuit. Over the years, various algorithms were proposed to tackle this problem, among of which we mention the thresholding algorithm \citep{Elad_Book} and its iterative variant \citep{Daubechies2004}. When handling natural signals, this model has been commonly used for modeling local patches extracted from the global data mainly due to the computational difficulties related to the task of learning the dictionary \citep{Elad2006,Dong2011,Mairal2014,Romano2015b,Sulam2015}. However, in recent years an alternative to this patch-based processing has emerged in the form of the Convolutional Sparse Coding (CSC) model \citep{Bristow2013,Kong2014,Wohlberg2014,Gu2015,Heide2015,Papyan2016_1,Papyan2016_2}. This circumvents the aforementioned limitations by imposing a special structure -- a union of banded and Circulant matrices -- on the dictionary involved. The traditional sparse model has been extensively studied over the past two decades \citep{Elad_Book,foucart2013mathematical}. More recently, the convolutional extension was extensively analyzed in \citep{Papyan2016_1,Papyan2016_2}, shedding light on its theoretical aspects and prospects of success.

In this work, by leveraging the recent study of CSC, we aim to provide a new perspective on CNN, leading to a clear and profound theoretical understanding of this scheme, along with new insights. Embarking from the classic CSC, our approach builds upon the observation that similar to the original signal, the representation vector itself also admits a convolutional sparse representation. As such, it can be modeled as a superposition of atoms, taken from a different convolutional dictionary. This rationale can be extended to several layers, leading to the definition of our proposed ML-CSC model. Building on the recent analysis of the CSC, we provide a theoretical study of this novel model and its associated pursuits, namely the layered thresholding algorithm and the layered basis pursuit (BP).

Our analysis reveals the relation between the CNN and the ML-CSC model, showing that \emph{the forward pass of the CNN is in fact identical to our proposed pursuit -- the layered thresholding algorithm}. This connection is of significant importance since it gives a clear mathematical meaning, objective and model to the CNN architecture, which in turn can be accompanied by guarantees for the success of the forward pass, studied via the layered thresholding algorithm. Specifically, we show that the forward pass is guaranteed to recover an estimate of the underlying representations of an input signal, assuming these are sparse in a local sense. Moreover, considering a setting where a norm-bounded noise is added to the signal, we show that such a mild corruption in the input results in a bounded perturbation in the output -- indicating the stability of the CNN in recovering the underlying representations. Lastly, we exploit the answers to the above questions in order to propose an alternative to the commonly used forward pass algorithm, which is tightly connected to both deconvolutional \citep{zeiler2010deconvolutional,pu2016deep} and recurrent networks \citep{bengio1994learning}, and also related to residual networks \citep{he2015deep}. The proposed alternative scheme is accompanied by a thorough theoretical study. Although this and the analysis presented throughout this work focus on CNN, we will show that they also hold for fully connected networks.

This paper is organized as follows. In Section \ref{Sec:Background} we review the basics of both the CNN and the Sparse-Land model. We then define the proposed ML-CSC model in Section \ref{Sec:ML-CSC}, together with its corresponding deep sparse coding problem. In Section \ref{Sec:Crux}, we aim to solve this using the layered thresholding algorithm, which is shown to be equivalent to the forward pass of the CNN. Next, having established the relevance of our model to CNN, we proceed to its analysis in Section \ref{Sec:TheoreticalStudy}. Standing on these theoretical grounds, we then propose in Section \ref{Sec:Future} a provably improved pursuit, termed the layered BP, accompanied by its theoretical analysis. We revisit the assumptions of our model in Section \ref{Sec:CloserLook}. First, in Section \ref{Sec:SparseDictionary} we link the double sparsity model to ours by assuming the dictionaries throughout the layers are sparse. Then, in Section \ref{Sec:Stride} we consider an idea typically employed in CNN, termed spatial-stride, showing its benefits from a simple theoretical perspective. Combining our insights from Section \ref{Sec:SparseDictionary} and \ref{Sec:Stride}, we move to an experimental phase by constructing a family of signals satisfying the assumptions of our model, which are then used in order to verify our theoretical results. Finally, in Section \ref{Sec:Conclusion} we conclude the contributions of this paper and present several future directions.

\section{Background} \label{Sec:Background}
This section is divided into two parts: The first is dedicated to providing a simple mathematical formulation of the CNN and the forward pass, while the second reviews the Sparse-Land model and its various extensions. Readers familiar with these two topics can skip directly to Section 3, which moves to serve the main contribution of this work.

\subsection{Deep Learning - Convolutional Neural Networks} \label{Sec:CNN}

\begin{figure}[t]
	\centering
	\begin{subfigure}{1\textwidth}
		\centering
		\includegraphics[width=0.75\textwidth]{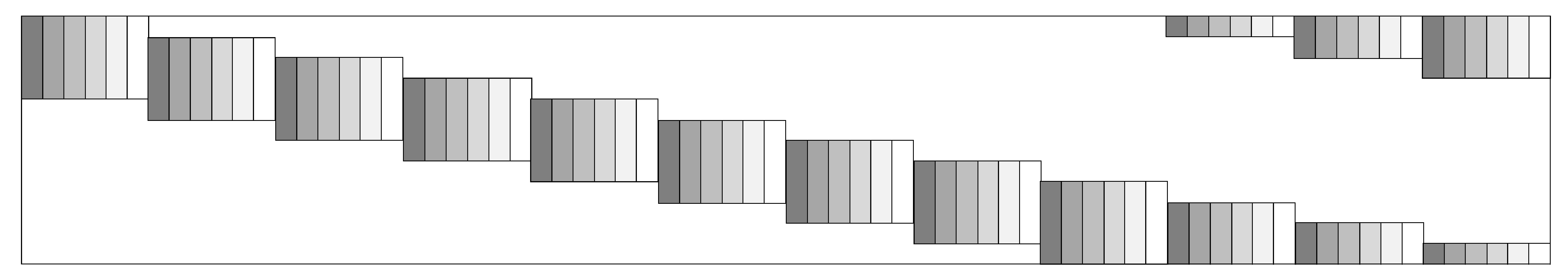}
		\caption{A convolutional matrix.}
		\label{Fig:Shifted_DL}
	\end{subfigure}
	\begin{subfigure}{1\textwidth}
		\centering
		\includegraphics[width=0.75\textwidth]{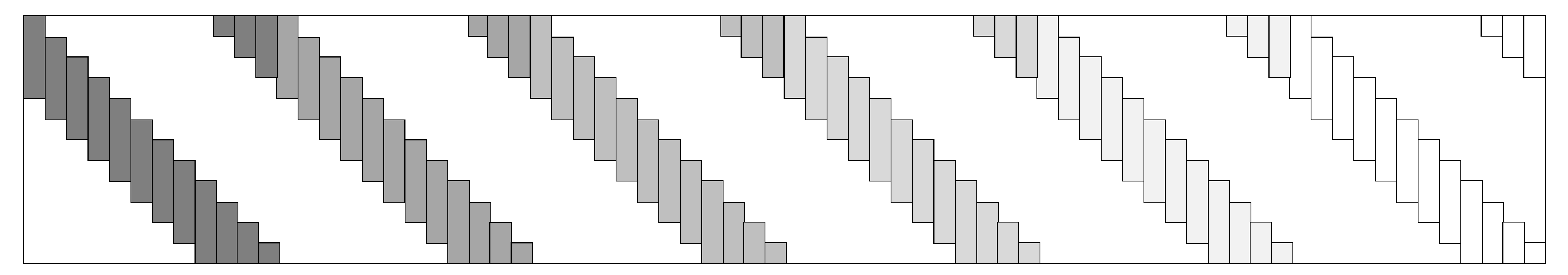}
		\caption{A concatenation of banded and Circulant matrices.}
		\label{Fig:CirculantDictionary}
	\end{subfigure}
	\caption{The two facets of the convolutional structure.}
\end{figure}

The fundamental algorithm of deep learning is the \emph{forward pass}, employed both in the training and the inference stages. The first step of this algorithm convolves an input (one dimensional) signal $\X\in\mathbb{R}^{N}$ with a set of $m_1$ learned filters of length $n_0$, creating $m_1$ feature (or kernel) maps. Equally, this convolution can be written as a matrix-vector multiplication, $\W_1^T\X\in\mathbb{R}^{N m_1}$, where $\W_1\in\mathbb{R}^{N\times N m_1}$ is a matrix containing in its columns the $m_1$ filters with all of their shifts. This structure, also known as a convolutional matrix, is depicted in Figure \ref{Fig:Shifted_DL}. A pointwise nonlinear function is then applied on the sum of the obtained feature maps $\W_1^T\X$ and a bias term denoted by $\b_1\in\mathbb{R}^{N m_1}$. Many possible functions were proposed over the years, the most popular one being the Rectifier Linear Unit (ReLU) \citep{glorot2011deep,krizhevsky2012imagenet}, formally defined as $\ReLU(z)=\max(z,0)$. By cascading the basic block of convolutions followed by a nonlinear function, $\Z_1 = \ReLU ( \W_1^T\X + \b_1 )$, a multi-layer structure of depth $K$ is constructed. Formally, for two layers this is given by
\begin{equation} \label{Eq:CNN}
f(\X,\{ \W_i \}_{i=1}^2,\{ \b_i \}_{i=1}^2) = \Z_2 = \ReLU \bigg( \ \W_2^T \ \ReLU \left( \ \W_1^T \X + \b_1 \ \right) + \b_2 \ \bigg),
\end{equation}
where $\W_2\in\mathbb{R}^{N m_1 \times N m_2}$ is a convolutional matrix (up to a small modification discussed below) constructed from $m_2$ filters of length $n_1m_1$ and $\b_2\in\mathbb{R}^{N m_2}$ is its corresponding bias. Although the two layers considered here can be readily extended to a much deeper configuration, we defer this to a later stage.

By changing the order of the columns in the convolutional matrix, one can observe that it can be equally viewed as a concatenation of banded and Circulant\footnote{We shall assume throughout this paper that boundaries are treated by a periodic continuation, which gives rise to the cyclic structure.} matrices, as depicted in Figure \ref{Fig:CirculantDictionary}. Using this observation, the above description for one dimensional signals can be extended to images, with the exception that now every Circulant matrix is replaced by a block Circulant with Circulant blocks one.

\begin{figure}[t]
	\centering
	\begin{subfigure}{1\textwidth}
		\centering
		\includegraphics[width=1\textwidth]{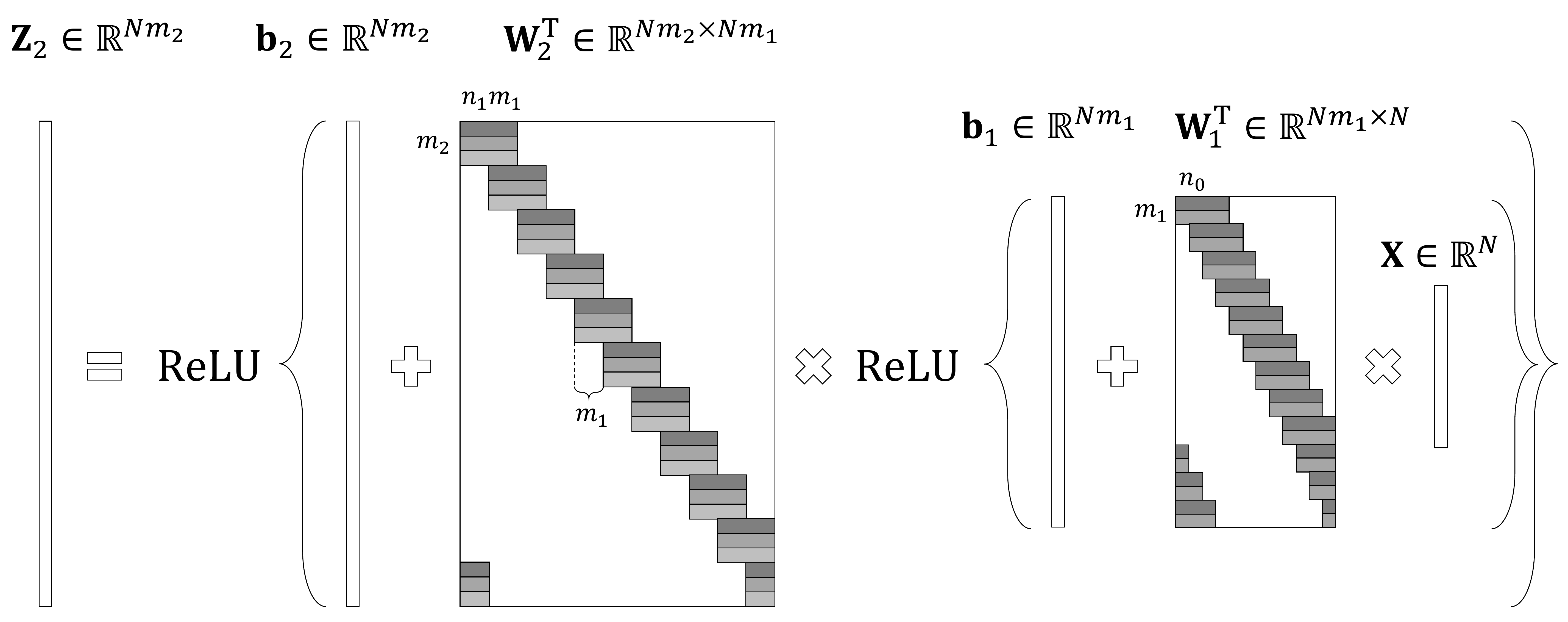}
		\caption{An illustration of Equation \eqref{Eq:CNN} for a one dimensional signal $\X$.}
		\label{Fig:ForwardPass1}
	\end{subfigure}
	\\[.35cm]
	\begin{subfigure}{0.9\textwidth}
		\centering
		\includegraphics[width=1\textwidth]{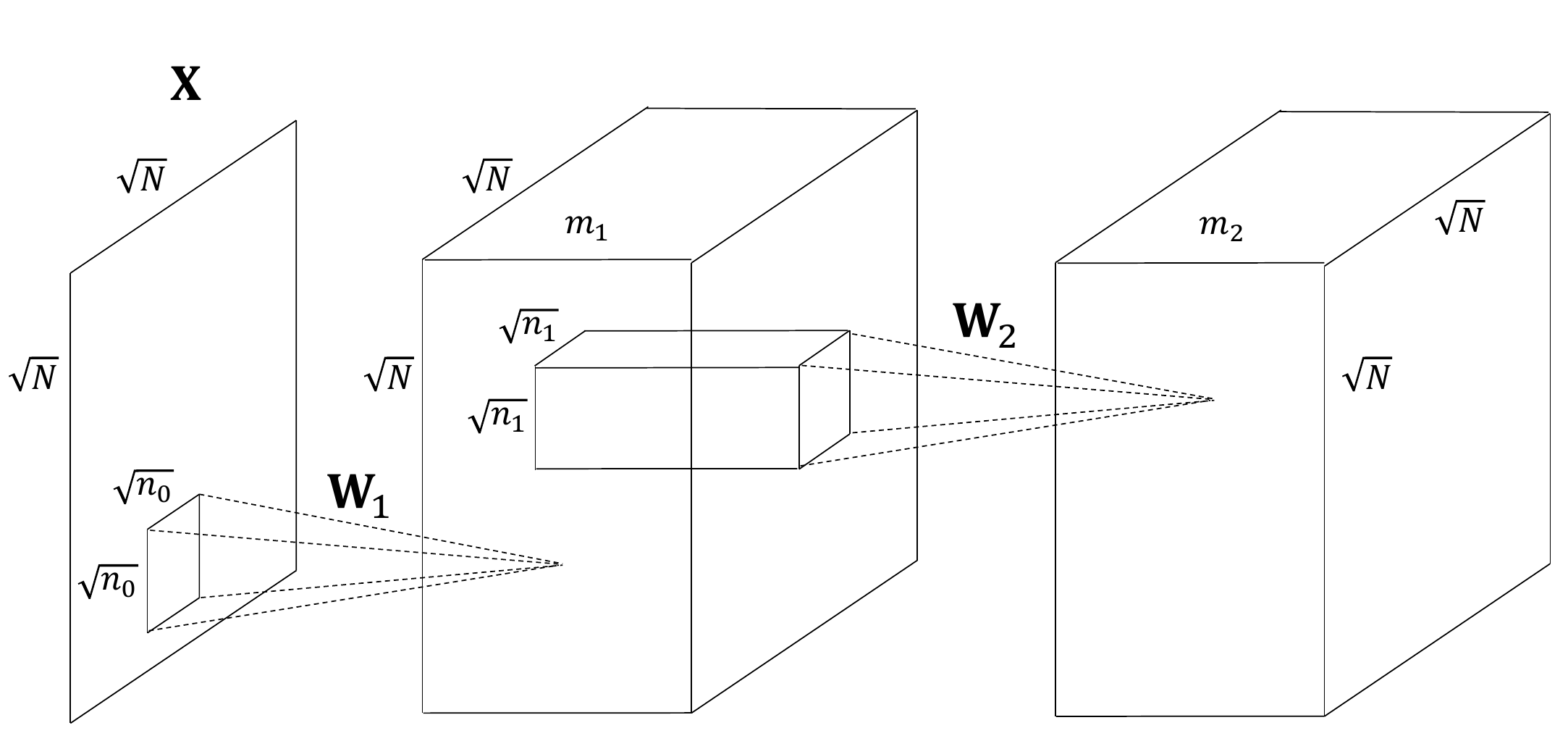}
		\caption{The evolution of an image $\X$ throughout the CNN. Notice that the number of channels in $\X$ is equal to one and as such $m_0=1$.}
		\label{Fig:ForwardPass2}
	\end{subfigure}
	\caption{The forward pass algorithm for a one dimensional signal (a) and an image (b).}
\end{figure}

An illustration of the forward pass algorithm is presented in Figure \ref{Fig:ForwardPass1} and \ref{Fig:ForwardPass2}. In Figure \ref{Fig:ForwardPass1} one can observe that $\W_2$ is not a regular convolutional matrix but a stride one, since it shifts local filters by skipping $m_1$ entries at a time. The reason for this becomes apparent once we look at Figure \ref{Fig:ForwardPass2}; the convolutions of the second layer are computed by shifting the filters of $\W_2$ that are of size $ \sqrt{n_1}\times\sqrt{n_1}\times m_1 $ across $N$ places, skipping $m_1$ indices at a time from the $\sqrt{N}\times\sqrt{N}\times m_1$-sized array. A matrix obeying this structure is called a \emph{stride convolutional matrix}.

Thus far, we have presented the basic structure of CNN. However, oftentimes an additional non-linear function, termed \emph{pooling}, is employed on the resulting feature map obtained from the ReLU operator. In essence, this step summarizes each $w_i$-dimensional spatial neighborhood from the $i$-th kernel map $\Z_i$ by replacing it with a single value. If the neighborhoods are non-overlapping, for example, this results in the down-sampling of the feature map by a factor of $w_i$. The most widely used variant of the above is the max pooling \citep{krizhevsky2012imagenet,simonyan2014very}, which picks the maximal value of each neighborhood. In \citep{springenberg2014striving} it was shown that this operator can be replaced by a convolutional layer with increased stride without loss in performance in several image classification tasks. Moreover, the current state-of-the-art in image recognition is obtained by the residual network \citep{he2015deep}, which does not employ any pooling steps (except for a single layer). As such, we defer the analysis of this operator to a follow-up work.

In the context of classification, for example, the output of the last layer is fed into a simple classifier that attempts to predict the label of the input signal $\X$, denoted by $h(\X)$. Given a set of signals $\{\X_j\}_j$, the task of learning the parameters of the CNN -- including the filters $\{ \W_i \}_{i=1}^K$, the biases $\{ \b_i \}_{i=1}^K$ and the parameters of the classifier $\U$ -- can be formulated as the following minimization problem
\begin{equation} \label{Eq:TrainingCNN}
\underset{\{ \W_i \}_{i=1}^K,\{ \b_i \}_{i=1}^K,\U}{\min} \ \sum_j \ell \Big( h(\X_j), \U, f\left( \X_j,\{ \W_i \}_{i=1}^K,\{ \b_i \}_{i=1}^K \right) \Big).
\end{equation}
This optimization task seeks for the set of parameters that minimize the mean of the loss function $\ell$, representing the price incurred when classifying the signal $\X$ incorrectly. The input for $\ell$ is the true label $h(\X)$ and the one estimated by employing the classifier defined by $\U$ on the final layer of the CNN given by $f\left( \X,\{ \W_i \}_{i=1}^K,\{ \b_i \}_{i=1}^K \right)$. Similarly one can tackle various other problems, e.g. regression or prediction.

In the remainder of this work we shall focus on the feature extraction process and assume that the parameters of the CNN model are pre-trained and fixed. These, for example, could have been obtained by minimizing the above objective via the backpropagation algorithm and the stochastic gradient descent, as in the VGG network \citep{simonyan2014very}.


\subsection{Sparse-Land}
This section presents an overview of the Sparse-Land model and its many extensions. We start with the traditional sparse representation and the core problem it aims to solve, and then proceed to its nonnegative variant. Next, we continue to the dictionary learning task both in the unsupervised and supervised cases. Finally, we describe the recent CSC model, which will lead us in the next section to the proposal of the ML-CSC model. This, in turn, will naturally connect the realm of sparsity to that of the CNN.

\subsubsection{Sparse Representation} \label{Sec:SparseRepresentation}
In the sparse representation model one assumes a signal $\X\in\mathbb{R}^{N}$ can be described as a multiplication of a matrix $\D\in\mathbb{R}^{N\times M}$, also called a dictionary, by a sparse vector $\Gama\in\mathbb{R}^M$. Equally, the signal $\X$ can be seen as a linear combination of a few columns from the dictionary $\D$, coined atoms.

For a fixed dictionary, given a signal $\X$, the task of recovering its sparsest representation $\Gama$ is called sparse coding, or simply pursuit, and it attempts to solve the following problem \citep{Donoho2003,Tropp2004,Elad_Book}:
\begin{equation}
(\Po): \ \underset{\Gama}{\min} \ \|\Gama\|_0 \ \text{ s.t. } \  \D\Gama = \X, \label{Eq:P0}
\end{equation}
where we have denoted by $\|\Gama\|_0$ the number of non-zeros in $\Gama$. The above has a convex relaxation in the form of the Basis-Pursuit (BP) problem \citep{Chen2001,Donoho2003,Tropp2006}, formally defined as
\begin{equation}
(\Pone): \ \underset{\Gama}{\min} \ \|\Gama\|_1 \ \text{ s.t. } \  \D\Gama = \X. \label{Eq:P1}
\end{equation}
Many questions arise from the above two defined problems. For instance, given a signal $\X$, is its sparsest representation unique? Assuming that such a unique solution exists, can it be recovered using practical algorithms such as the Orthogonal Matching Pursuit (OMP) \citep{Chen1989,Pati1993a} and the BP \citep{Chen2001,Daubechies2004}? The answers to these questions were shown to be positive under the assumption that the number of non-zeros in the underlying representation is not too high and in particular less than $\frac{1}{2}\left( 1 + \frac{1}{\mu(\D)} \right)$ \citep{Donoho2003,Tropp2004,Donoho2006}. The quantity $\mu(\D)$ is the mutual coherence of the dictionary $\D$, being the maximal inner product of two atoms extracted from it\footnote{Hereafter, we assume that the atoms are normalized to a unit $\ell_2$ norm.}. Formally, we can write
\begin{equation}
\mu(\D) = \underset{i\neq j}{\max} \ |\d_i^T \d_j|.
\end{equation}
Tighter conditions, relying on sharper characterizations of the dictionary, were also suggested in the literature \citep{candes2006stable,schnass2007average,candes2006stable,candes2007dantzig}. However, at this point, we shall not dwell on these.

One of the simplest approaches for tackling the $\Po$ and $\Pone$ problems is via the hard and soft thresholding algorithms, respectively. These operate by computing the inner products between the signal $\X$ and all the atoms in $\D$ and then choosing the atoms corresponding to the highest responses. This can be described as solving, for some scalar $\beta$, the following problems:
\begin{equation}
\underset{\Gama}{\min} \ \frac{1}{2}\| \Gama - \D^T\X \|_2^2 + \beta\|\Gama\|_0
\end{equation}
for the $\Po$, or
\begin{equation} \label{Eq:L1Projection}
\underset{\Gama}{\min} \ \frac{1}{2}\| \Gama - \D^T\X \|_2^2 + \beta\|\Gama\|_1,
\end{equation}
for the $\Pone$. The above are simple projection problems that admit a closed-form solution in the form\footnote{The curious reader may identify the relation between the notations used here and the ones in the previous subsection, which starts to reveal the relation between CNN and sparsity-inspired models. This connection will be made stringer and clearer as we proceed to CSC.} of $\H_\beta(\D^T\X)$ or $\S_\beta(\D^T\X)$, where we have defined the hard thresholding operator $\H_{\beta}(\cdot)$ by
\begin{equation}
\H_\beta(z)= 
\begin{cases}
z, & z < -\beta \\
0, & -\beta\leq z\leq \beta \\
z, & \beta < z,
\end{cases}
\end{equation}
and the soft thresholding operator $\S_{\beta}(\cdot)$ by
\begin{equation} \label{Eq:SoftThresholding}
    \S_\beta(z)= 
    \begin{cases}
	    z+\beta, & z < -\beta \\
	    0, & -\beta\leq z\leq \beta \\
	    z-\beta, & \beta < z.
	\end{cases}
\end{equation}
Both of the above, depicted in Figure \ref{Fig:Thresholding_Plots}, nullify small entries and thus promote a sparse solution. However, while the hard thresholding operator does not modify large coefficients (in absolute value), the soft thresholding does, by contracting these to zero. This inherent limitation of the soft version will appear later on in our theoretical analysis.

As for the theoretical guarantees for the success of the simple thresholding algorithms; these depend on the properties of $\D$ and on the ratio between the minimal and maximal coefficients in absolute value in $\Gama$, and thus are weaker when compared to those found for OMP and BP \citep{Donoho2003,Tropp2004,Donoho2006}. Still, under some conditions, both algorithms are guaranteed to find the true support of $\Gama$ along with an approximation of its true coefficients. Moreover, a better estimation of these can be obtained by projecting the input signal onto the atoms corresponding to the found support (indices of the non-zero entries) by solving a Least-Squares problem. This step, termed debiasing \citep{Elad_Book}, results in a more accurate identification of the non-zero values.

\subsubsection{Nonnegative Sparse Coding} \label{Sec:Nonnegative}
The nonnegative sparse representation model assumes a signal can be decomposed into a multiplication of a dictionary and a \emph{nonnegative} sparse vector. A natural question arising from this is whether such a modification to the original Sparse-Land model affects its expressiveness. To address this, we hereby provide a simple reduction from the original sparse representation to the nonnegative one.

Consider a signal $\X=\D\Gama$, where the signs of the entries in $\Gama$ are unrestricted. Notice that this can be equally written as
\begin{equation}
\X = \D\Gama_P + (-\D)(-\Gama_N),
\end{equation}
where we have split the vector $\Gama$ to its positive coefficients, $\Gama_P$, and its negative ones, $\Gama_N$. Since the coefficients in $\Gama_P$ and $-\Gama_N$ are all positive, one can thus assume the signal $\X$ admits a non-negative sparse representation over the dictionary $\left[ \D, -\D \right]$ with the vector $\left[ \Gama_P, -\Gama_N \right]^T$. Thus, restricting the coefficients in the sparsity inspired model to be nonnegative does not change its expressiveness.

\begin{figure}[t]
	\centering
	\setlength\figurewidth{0.6\textwidth}
	\setlength\figureheight{0.4\textwidth}
	\input{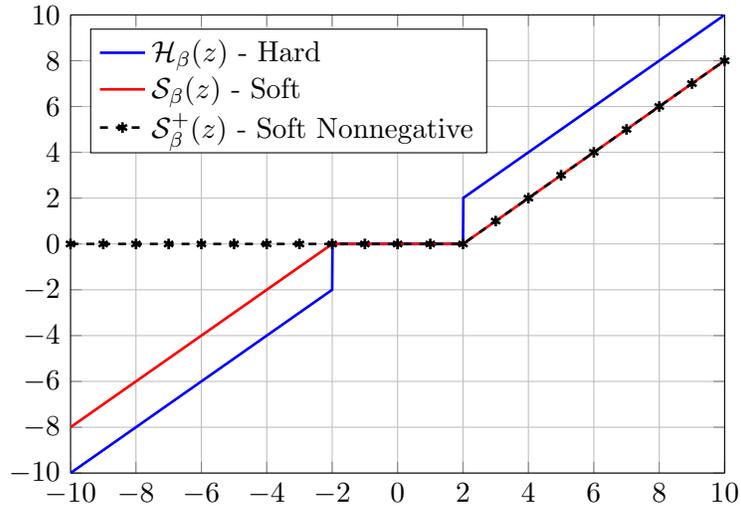}
	\caption{The thresholding operators for a constant $\beta=2$.}
	\label{Fig:Thresholding_Plots}
\end{figure}

Similar to the original model, in the nonnegative case, one could solve the associated pursuit problem by employing a soft thresholding algorithm. However, in this case a constraint must be added to the optimization problem in Equation \eqref{Eq:L1Projection}, forcing the outcome to be positive, i.e.,
\begin{equation}
\underset{\Gama}{\min} \ \frac{1}{2}\| \Gama - \D^T\X \|_2^2 + \beta\|\Gama\|_1 \quad \text{s.t.} \quad \Gama\geq \mathbf{0}.
\end{equation}
Since the above is a simple projection problem (onto the $\ell_1$ ball constrained to positive entries), it admits a closed-form solution $\S_\beta^+(\D^T\X)$, where we have defined the soft nonnegative thresholding operator $\S_\beta^+(\cdot)$ as
\begin{equation} \label{Eq:SoftNonnegativeThresholding}
    \S_\beta^+(z)= 
    \begin{cases}
	    0, & z\leq \beta \\
	    z-\beta, & \beta<z.
	\end{cases}
\end{equation}
Remarkably, the above function satisfies
\begin{equation} \label{Eq:ReLU_SoftNonnegativeThresholding}
\S_\beta^+(z)=\max(z-\beta,0)=\ReLU(z-\beta).
\end{equation}
In other words, the ReLU and the soft nonnegative thresholding operator are equal, a fact that will prove to be important later in our work. We should note that a similar conclusion was reached in \citep{fawzi2015dictionary}. To summarize this discussion, we depict in Figure \ref{Fig:Thresholding_Plots} the hard, soft, and nonnegative soft thresholding operators.

\subsubsection{Unsupervised and Task Driven Dictionary Learning} \label{Sec:DictionaryLearning}
At first, the dictionaries employed in conjunction with the sparsity inspired model were analytically defined matrices, such as the Wavelet and the Fourier \citep{daubechies1992ten,Mallat1993,elad2002generalized,Mallat_AwaveletTour}. Although the sparse coding problem under these can be done very efficiently, over the years many have shifted to a data driven approach -- adapting the dictionary $\D$ to a set of training signals at hand via some learning procedure. This was empirically shown to lead to sparser representations and better overall performance, at the cost of complicating the involved pursuit, since the dictionary was usually chosen to be redundant (having more columns than rows).

The task of learning a dictionary for representing a set of signals $\{\X_j\}_j$ can be formulated as follows
\begin{equation}
\underset{\D, \{\Gama^j\}_j}{\min} \sum_j \| \X_j - \D\Gama^j \|_2^2 + \xi\| \Gama^j \|_0.
\end{equation}
The above formulation is an unsupervised learning procedure, and it was later extended to a supervised setting. In this context, given a set of signals $\{ \X_j \}_j$, one attempts to predict their corresponding labels $\{ h(\X_j) \}_j$. A common approach for tackling this is first solving a pursuit problem for each signal $\X_j$ over a dictionary $\D$, resulting in
\begin{equation}
\Gama^{\star}(\X_j,\D) = \underset{\Gama}{\arg\min} \ \| \Gama \|_0 \quad \text{s.t.} \quad \D\Gama = \X_j,
\end{equation}
and then feeding these sparse representations into a simple classifier, defined by the parameters $\U$. The task of learning jointly the dictionary $\D$ and the classifier $\U$ was addressed in \citep{mairal2012task}, where the following optimization problem was proposed
\begin{equation}
\underset{\D,\U}{\min} \ \sum_j \ell \Big( h(\X_j),\U,\Gama^{\star}(\X_j,\D) \Big).
\end{equation}
The loss function $\ell$ in the above objective penalizes the estimated label if it is different from the true $h(\X_j)$, similar to what we have seen in Section \ref{Sec:CNN}. The above formulation contains in it the unsupervised option as a special case, in which $\U$ is of no importance, and the loss function is the representation error $\sum_j \| \X_j - \D \Gama^{\star}_j \|_2^2$.

Double sparsity -- first proposed in \citep{Rubinstein2010} and later employed in \citep{sulam2016trainlets} -- attempts to benefit from both the computational efficiency of analytically defined matrices, and the adaptability of data driven dictionaries. In this model, one assumes the dictionary $\D$ can be factorized into a multiplication of two matrices, $\D_1$ and $\D_2$, where $\D_1$ is an analytic dictionary with fast implementation, and $\D_2$ is a trained sparse one. As a result, the signal $ \X $ can be represented as 
\begin{equation}
\X = \D\Gama_2 = \D_1\D_2\Gama_2,
\end{equation}
where $\Gama_2$ is sparse.

We propose a different interpretation for the above, which is unrelated to practical aspects. Since both the matrix $\D_2$ and the vector $\Gama_2$ are sparse, one would expect their multiplication $\Gama_1=\D_2\Gama_2$ to be sparse as well. As such, the double sparsity model implicitly assumes that the signal $\X$ can be decomposed into a multiplication of a dictionary $\D_1$ and sparse vector $\Gama_1$, which in turn can also be decomposed similarly via $\Gama_1=\D_2\Gama_2$.

\subsubsection{Convolutional Sparse Coding Model}
Due to the computational constraints entailed when deploying trained dictionaries, this approach seems valid only for treatment of low-dimensional signals. Indeed, the sparse representation model is traditionally used for modeling local patches extracted from a global signal. An alternative, which was recently proposed, is the CSC model that attempts to represent the whole signal $\X\in\mathbb{R}^{N}$ as a multiplication of a global convolutional dictionary $\D\in\mathbb{R}^{N\times Nm}$ and a sparse vector $\Gama\in\mathbb{R}^{Nm}$. Interestingly, the former is constructed by shifting a local matrix of size $n\times m$ in all possible positions, resulting in the same structure as the one shown in Figure \ref{Fig:Shifted_DL}.

In the convolutional model, the classical theoretical guarantees (we are referring to results reported in \citep{Chen2001,Donoho2003,Tropp2006}) for the $\Po$ problem, defined in Equation \eqref{Eq:P0}, are very pessimistic. In particular, the condition for the uniqueness of the underlying solution and the requirement for the success of the sparse coding algorithms depend on the global number of non-zeros being less than $\frac{1}{2}\left( 1 + \frac{1}{\mu(\D)} \right)$. Following the Welch bound \citep{Welch1974}, this expression was shown in \citep{Papyan2016_1} to be impractical, allowing the global number of non-zeros in $\Gama$ to be extremely low.

\begin{figure}[t]
	\centering
	\includegraphics[width=1\textwidth]{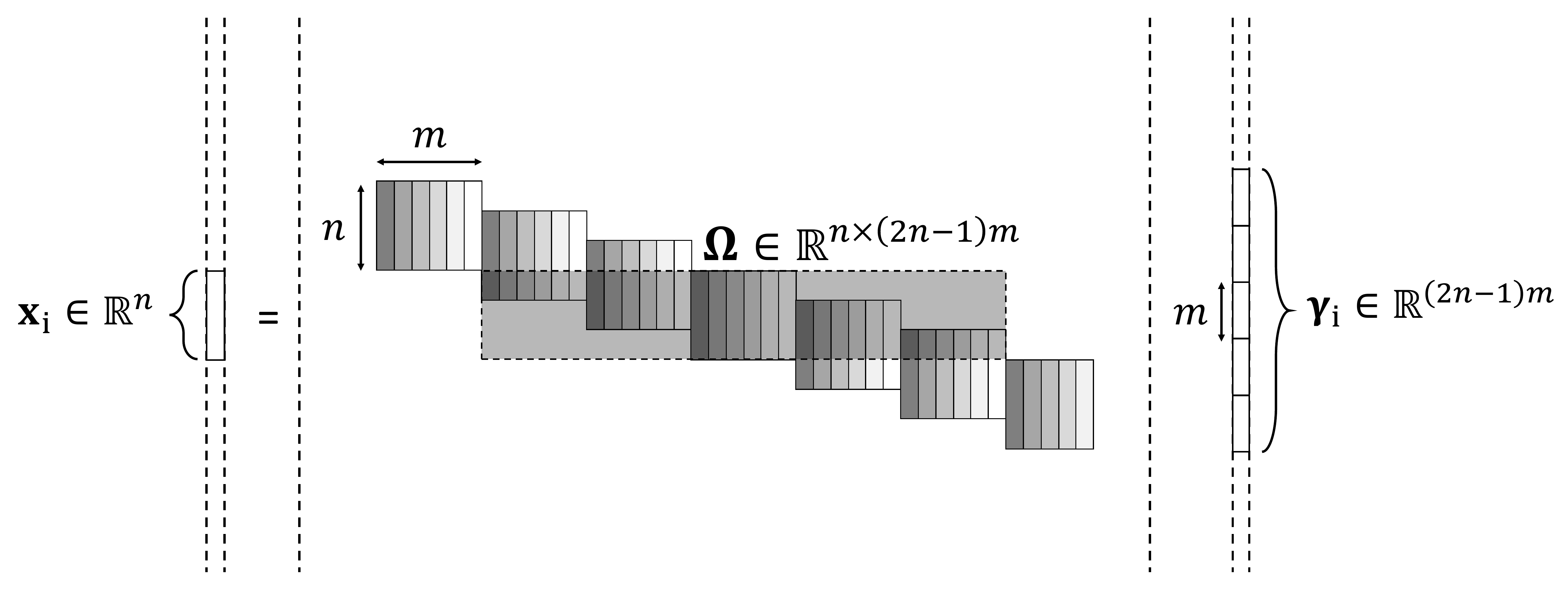}
	\caption{The $i$-th patch $\x_i$ of the global system $\X=\D\Gama$, given by $\x_i=\O\gama_i$.}
	\label{Fig:LocalSystem}
\end{figure}

In order to provide a better theoretical understanding of this model, which exploits the inherent structure of the convolutional dictionary, a recent work \citep{Papyan2016_1} suggested to measure the sparsity of $\Gama$ in a localized manner. More concretely, consider the $i$-th $n$-dimensional patch of the global system $\X=\D\Gama$, given by $\x_i=\O\gama_i$. The stripe-dictionary $\O$, which is of size $n\times(2n-1)m$, is obtained by extracting the $i$-th patch from the global dictionary $\D$ and discarding all the zero columns from it. The stripe vector $\gama_i$ is the corresponding sparse representation of length $(2n-1)m$, containing all coefficients of atoms contributing to $\x_i$. This relation is illustrated in Figure \ref{Fig:LocalSystem}. Notably, the choice of a convolutional dictionary results in signals such that every patch of length $n$ extracted from them can be sparsely represented using a single shift-invariant local dictionary $\O$ -- a common assumption usually employed in signal and image processing.

Following the above construction, the $\Loi$ norm of the global sparse vector $\Gama$ is defined to be the maximal number of non-zeros in a stripe of length $(2n - 1)m$ extracted from it. Formally,
\begin{equation}
\|\Gama\|_{0,\infty}^\ss = \max_i \|\gama_i\|_0,
\end{equation}
where the letter \textbf{s} emphasizes that the $\Loi$ norm is computed by sweeping over all stripes. Given a signal $\X$, finding its sparest representation $\Gama$ in the $\Loi$ sense is equal to the following optimization problem:
\begin{equation} \label{Eq:Poi}
(\Poi): \quad \min_\Gama \quad \|\Gama\|_{0,\infty}^\ss \ \text{ s.t. }\ \D\Gama=\X.
\end{equation}
Intuitively, this seeks for a global vector $\Gama$ that can represent sparsely every patch in the signal $\X$ using the dictionary $\O$. The advantage of the above problem over the traditional $\Po$ becomes apparent as we move to consider its theoretical aspects. Assuming that the \textbf{number of non-zeros per stripe} (and not globally) in $\Gama$ is less than $\frac{1}{2}\left( 1 + \frac{1}{\mu(\D)} \right)$, in \citep{Papyan2016_1} it was proven that the solution for the $\Poi$ problem is unique. Furthermore, classical pursuit methods, originally tackling the $\Po$ problem, are guaranteed to find this representation.

When modeling natural signals, due to measurement noise as well as model deviations, one can not impose a perfect reconstruction such as $\X=\D\Gama$ on the signal $\X$. Instead, one assumes $\Y = \X + \E = \D \Gama + \E$, where $\E$ is, for example, an $\ell_2$-bounded error vector. To address this, the work reported in \citep{Papyan2016_2} considered the extension of the $\Poi$ problem to the $\Poie$ one, formally defined as
\begin{equation}
\quad (\Poie): \quad \underset{\Gama}{\min} \ \|\Gama\|_{0,\infty}^\ss \ \text{ s.t. }\ \|\Y-\D\Gama\|^2_2\leq\Eps^2.
\end{equation}
Similar to the $\Poi$ problem, this was also analyzed theoretically, shedding light on the theoretical aspects of the convolutional model in the presence of noise. In particular, a stability claim for the $\Poie$ problem and guarantees for the success of both the OMP and the BP were provided. Similar to the noiseless case, these assumed that the number of non-zeros per stripe is low.

\section{From Atoms to Molecules: Multi-Layer Convolutional Sparse Model} \label{Sec:ML-CSC}
Convolutional sparsity assumes an inherent structure for natural signals. Similarly, the representations themselves could also be assumed to have such a structure. In what follows, we propose a novel layered model that relies on this rationale.

The convolutional sparse model assumes a global signal $\X\in\mathbb{R}^{N}$ can be decomposed into a multiplication of a convolutional dictionary $\D_1\in\mathbb{R}^{N\times Nm_1}$, composed of $m_1$ local filters of length $n_0$, and a sparse vector $\Gama_1\in\mathbb{R}^{Nm_1}$. Herein, we extend this by proposing a similar factorization of the vector $\Gama_1$, which can be perceived as an $N$-dimensional global signal with $m_1$ channels. In particular, we assume $\Gama_1=\D_2\Gama_2$, where $\D_2\in\mathbb{R}^{Nm_1\times Nm_2}$ is a stride convolutional dictionary (skipping $m_1$ entries at a time) and $\Gama_2\in\mathbb{R}^{Nm_2}$ is a sparse representation. We denote the number of unique filters constructing $\D_2$ by $m_2$ and their corresponding length by $n_1 m_1$. Due to the multi-layer nature of this model and the imposed convolutional structure, we name this the ML-CSC model.

Intuitively, $\X = \D_1\Gama_1$ assumes that the signal $\X$ is a superposition of \textbf{atoms} taken from $\D_1$. While equation $\X = \D_1\D_2\Gama_2$ views the signal as a superposition of more complex entities taken from the dictionary $\D_1\D_2$, which we coin \textbf{molecules}.

While this proposal can be interpreted as a straightforward fusion between the double sparsity model \citep{Rubinstein2010} and the convolutional one, it is in fact substantially different. The double sparsity model assumes that $\D_2$ is sparse, and forces \textbf{only} the deepest representation $\Gama_2$ to be sparse as well. Here, on the other hand, we replace this constraint by forcing $\D_2$ to have a stride convolution structure, putting emphasis on the sparsity of both the representations $\Gama_1$ and $\Gama_2$. In Section \ref{Sec:SparseDictionary} we will revisit the double sparsity work and its ties to ours by showing the benefits of injecting the assumption on the sparsity of $\D_2$ into our proposed model.

\begin{figure}[t]
	\centering
	\includegraphics[width=0.9\textwidth]{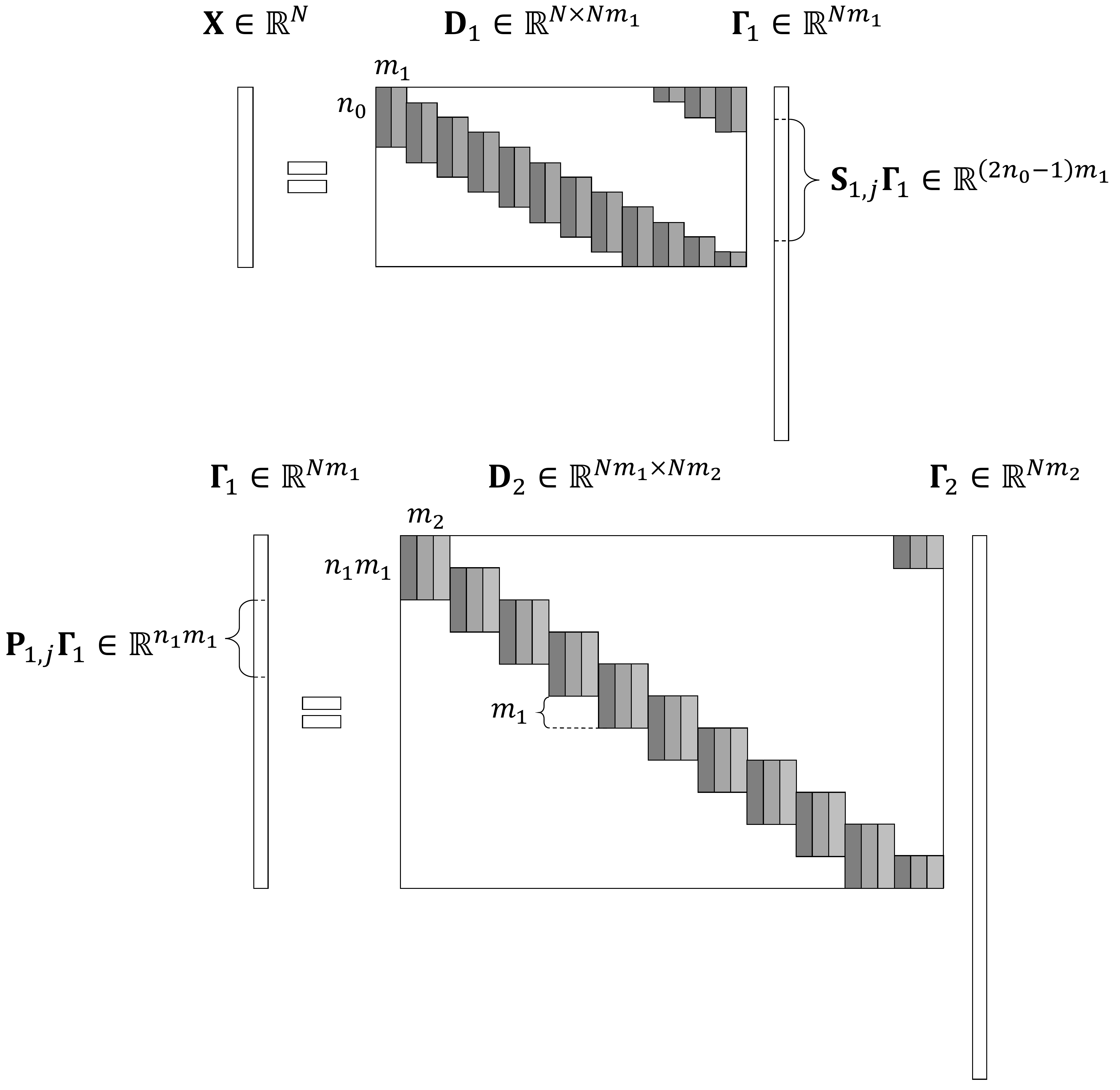}
	\caption{An instance $\X = \D_1\Gama_1 = \D_1\D_2\Gama_2$ of the ML-CSC model. Notice that $\Gama_1$ is built of both stripes $\SS_{1,j}\Gama_1$ and patches $\PP_{1,j}\Gama_1$.}
	\label{Fig:GlobalSystem}
\end{figure}

Under the above construction the sparse vector $\Gama_1$ has two roles. In the context of the system of equations $\X = \D_1\Gama_1$, it is the convolutional sparse representation of the signal $\X$ over the dictionary $\D_1$. As such, the vector $\Gama_1$ is composed from $(2n_0 -1 )m_1$-dimensional stripes, $\SS_{1,j}\Gama_1$, where $\SS_{i,j}$ is the operator that extracts the $j$-th stripe from $\Gama_i$. From another point of view, $\Gama_1$ is in itself a signal that admits a sparse representation $\Gama_1 = \D_2\Gama_2$. Denoting by $\PP_{i,j}$ the operator that extracts the $j$-th patch from $\Gama_i$, the signal $\Gama_1$ is composed of patches $\PP_{1,j}\Gama_1$ of length $n_1 m_1$. The above model is depicted in Figure \ref{Fig:GlobalSystem}, presenting both roles of $\Gama_1$ and their corresponding constituents -- stripes and patches. Clearly, the above construction can be extended to more than two layers, leading to the following definition:
\begin{definition} \label{Def:DCP}
	For a global signal $\X$, a set of convolutional dictionaries $\{ \D_i \}_{i=1}^K$, and a vector $\lamda$, define the deep coding problem $\DCP$ as:
	\begin{align*}
	\hspace{1.5cm} (\DCP): \quad \text{find} \quad \{\Gama_i\}_{i=1}^{K} \qquad \text{ s.t. } \qquad
	\X           & = \D_1 \Gama_1,       & \| \Gama_1 \|_{0,\infty}^\ss		& \leq \lambda_1 \hspace{4cm} \\
	\Gama_1      & = \D_2 \Gama_2,       & \| \Gama_2 \|_{0,\infty}^\ss		& \leq \lambda_2 \\
	             & \phantom{..} \vdots   &									& \phantom{..} \vdots \\
	\Gama_{K-1}  & = \D_K \Gama_K,       & \| \Gama_{K} \|_{0,\infty}^\ss	& \leq \lambda_K,
	\end{align*}
	where the scalar $\lambda_i$ is the $i$-th entry of $\lamda$.
\end{definition}

\noindent
Denoting $\Gama_0$ to be the signal $\X$, the $\DCP$ can be rewritten compactly as
\begin{equation}
(\DCP): \quad \text{find} \quad \{\Gama_i\}_{i=1}^{K} \quad \text{ s.t. } \quad \Gama_{i-1} = \D_i \Gama_i, \quad \| \Gama_i \|_{0,\infty}^\ss \leq \lambda_i, \quad \forall 1\leq i\leq K.
\end{equation}
Intuitively, given a signal $\X$, this problem seeks for a set of representations, $\{\Gama_i\}_{i=1}^{K}$, such that each one is locally sparse. As we shall see next, the above can be easily solved using simple algorithms that also enjoy from theoretical justifications. Next, we extend the $\DCP$ problem to a noisy regime.
\begin{definition}
	For a global signal $\Y$, a set of convolutional dictionaries $\{ \D_i \}_{i=1}^K$, and vectors $\lamda$ and $\Eps$, define the deep coding problem $\DCPE$ as:
	\begin{align*}
	\hspace{1cm} (\DCPE): \quad \text{find} \quad \{\Gama_i\}_{i=1}^{K} \qquad \text{ s.t. } \qquad
	\| \Y - \D_1 \Gama_1\|_2 & \leq \mathcal{E}_0, & \| \Gama_1 \|_{0,\infty}^\ss & \leq \lambda_1 \\
	\| \Gama_1 - \D_2 \Gama_2 \|_2 & \leq \mathcal{E}_1, & \| \Gama_2 \|_{0,\infty}^\ss & \leq \lambda_2 \\
	& \phantom{..} \vdots  & & \phantom{..} \vdots \hspace{3cm} \\
	\| \Gama_{K-1} - \D_K \Gama_K \|_2 & \leq \mathcal{E}_{K-1}, & \| \Gama_{K} \|_{0,\infty}^\ss & \leq \lambda_K,
	\end{align*}
	where the scalars $\lambda_i$ and $\mathcal{E}_i$ are the $i$-th entry of $\lamda$ and $\Eps$, respectively.
\end{definition}

We now move to the task of learning the model parameters. Denote by $\DCPstar(\X,\{ \D_i \}_{i=1}^K)$ the representation $\Gama_K$ obtained by solving the DCP problem (Definition \ref{Def:DCP}, i.e., noiseless) for the signal $\X$ and the set of dictionaries $\{ \D_i \}_{i=1}^K$. Relying on this, we now extend the dictionary learning problem, as presented in Section \ref{Sec:DictionaryLearning}, to the multi-layer convolutional sparse representation setting.
\begin{definition}
	For a set of global signals $\{ \X_j \}_j$, their corresponding labels $\{ h(\X_j) \}_j$, a loss function $\ell$, and a vector $\lamda$, define the deep learning problem $\DLP$ as:
	\begin{equation} \label{Eq:DLP}
	(\DLP): \quad \underset{\{ \D_i \}_{i=1}^K,\U}{\min} \ \sum_j \ell \Big( h(\X_j),\U,\DCPstar(\X_j,\{ \D_i \}_{i=1}^K) \Big).
	\end{equation}
\end{definition}
\noindent
A clarification for the chosen name, deep learning problem, will be provided shortly. The solution for the above results in an end-to-end mapping, from a set of input signals to their corresponding labels. Similarly, we can define the $\DLPE$ problem. However, this is omitted for the sake of brevity. We conclude this section by summarizing, for the convenience of the reader, all notations used throughout this work in Table \ref{Table:Notations}.

\begin{table}[t]
	\centering
	\begin{tabulary}{1.0\textwidth}{|p{1.9cm}@{:\quad}p{12.5cm}|} \hline
		$\X = \Gama_0$					& a global signal of length $N$. \\ \hline
		$\E$, $\Y = \hat{\Gama}_0$		& a global error vector and its corresponding noisy signal, where generally $\Y=\X+\E$. \\ \hline
		$K$								& the number of layers. \\ \hline
		$m_i$							& the number of local filters in $\D_i$, and also the number of channels in $\Gama_i$. Notice that $m_0 = 1$. \\ \hline	
		$n_0$							& the size of a local patch in $\X = \Gama_0$. \\ \hline
		$n_i, \ i\geq 1$				& the size of a local patch (not including channels) in $\Gama_i$. \\ \hline
		$n_i m_i$						& the size of a local patch (including channels) in $\Gama_i$. \\ \hline		
		$\D_1$							& a (full) convolutional dictionary of size $N\times N m_1 $ with filters of length $n_0$. \\ \hline
		$\D_i, \ i\geq 2$				& a convolutional dictionary of size $N m_{i-1}\times N m_i$ with filters of length $n_{i-1}m_{i-1}$ and a stride equal to $m_{i-1}$. \\ \hline	
		$\Gama_i$						& a sparse vector of length $N m_i$ that is the representation of $\Gama_{i-1}$ over the dictionary $\D_i$, i.e. $\Gama_{i-1} = \D_i \Gama_i$. \\ \hline
		$\SS_{i,j}$						& an operator that extracts the $j$-th stripe of length $(2 n_{i-1} - 1)m_i$ from $\Gama_i$. \\ \hline
		$\|\Gama_i\|_{0,\infty}^\ss$	& the maximal number of non-zeros in a stripe from $\Gama_i$. \\ \hline
		$\PP_{i,j}$						& an operator that extracts the $j$-th $n_i m_i$-dimensional patch from $\Gama_i$. \\ \hline
		$\|\Gama_i\|_{0,\infty}^\pp$	& the maximal number of non-zeros in a patch from $\Gama_i$ (Definition \ref{Def:LtinfLoip}). \\ \hline
		$\RR_{i,j}$						& an operator that extracts the filter of length $n_{i-1} m_{i-1}$ from the $j$-th atom in $\D_i$. \\ \hline
		$\|\V\|_{2,\infty}^\pp$			& the maximal $\ell_2$ norm of a patch extracted from a vector $\V$ (Definition \ref{Def:LtinfLoip}). \\ \hline
	\end{tabulary}
	\caption{Summary of notations used throughout the paper.}
	\label{Table:Notations}
	\vspace{-0.4cm}
\end{table}

\section{Layered Thresholding: The Crux of the Matter} \label{Sec:Crux}
Consider the ML-CSC model defined by the set of dictionaries $\{ \D_i \}_{i=1}^K$. Assume we are given a signal
\begin{align}
\X & = \D_1\Gama_1 \\
\Gama_1 & = \D_2\Gama_2 \\
& \phantom{..} \vdots \\
\Gama_{K-1} & = \D_K\Gama_K,
\end{align}
and our goal is to find its underlying representations, $\{\Gama_i\}_{i=1}^{K}$. Tackling this problem by recovering all the vectors at once might be computationally and conceptually challenging; therefore, we propose the \emph{layered thresholding algorithm} that gradually computes the sparse vectors one at a time across the different layers. Denoting by $\P_\beta(\cdot)$ a sparsifying operator that is equal to $\H_\beta(\cdot)$ in the hard thresholding case and $\S_\beta(\cdot)$ in the soft one; we commence by computing $\hat{\Gama}_1 = \P_{\beta_1}(\D_1^T\X)$, which is an approximation of $\Gama_1$. Next, by applying another thresholding algorithm, however this time on $\hat{\Gama}_1$, an approximation of $\Gama_2$ is obtained, $\hat{\Gama}_2 = \P_{\beta_2}(\D_2^T\hat{\Gama}_1)$. This process, which is iterated until the last representation $\hat{\Gama}_K$ is acquired, is summarized in Algorithm \ref{Alg:LayeredThresholding}.

One might ponder as to why does the application of the thresholding algorithm on the signal $\X$ not result in the true representation $\Gama_1$, but instead an approximation of it. As previously described in Section \ref{Sec:SparseRepresentation}, assuming some conditions are met, the result of the thresholding algorithm, $\hat{\Gama}_1$, is guaranteed to have the correct support. In order to obtain the vector $\Gama_1$ itself, one should project the signal $\X$ onto this obtained support, by solving a Least-Squares problem. For reasons that will become clear shortly, we choose not to employ this step in the layered thresholding algorithm. Despite this algorithm failing in recovering the exact representations in the noiseless setting, as we shall see in Section \ref{Sec:TheoreticalStudy}, the estimated sparse vectors and the true ones are close -- indicating the stability of this simple algorithm.

Thus far, we have assumed a noiseless setting. However, the same layered thresholding algorithm could be employed for the recovery of the representations of a noisy signal $\Y = \X + \E$, with the exception that the threshold constants, $\{\beta_i\}_{i=1}^{K}$, would be different and proportional to the noise level.

\begin{algorithm}[t]
	\caption{The layered thresholding algorithm.}
	\label{Alg:LayeredThresholding}
	
	\textbf{Input:}
	\begin{algorithmic}
	\State $\X$ -- a signal.
	\State $\{ \D_i \}_{i=1}^K$ -- convolutional dictionaries.
	\State $\P \in \{ \H, \S, \S^+ \}$ -- a thresholding operator.
	\State $\{\beta_i\}_{i=1}^{K}$ -- thresholds.
	\end{algorithmic}
	
	\textbf{Output:}
	\begin{algorithmic}
	\State A set of representations $\{\hat{\Gama}_i\}_{i=1}^{K}$.
	\end{algorithmic}

	\textbf{Process:}
	\begin{algorithmic}[1]
	\State $\hat{\Gama}_0 \leftarrow \X$
	
	\For{$i=1:K$}
	\State $\hat{\Gama}_i \leftarrow \P_{\beta_i}(\D_i^T\hat{\Gama}_{i-1})$
	\EndFor
	\end{algorithmic}
\end{algorithm}

Assuming two layers for simplicity, Algorithm \ref{Alg:LayeredThresholding} can be summarized in the following equation
\begin{equation}
\hat{\Gama}_2 = \P_{\beta_2} \bigg( \ \D_2^T \P_{\beta_1} \left( \ \D_1^T \X \ \right) \ \bigg).
\end{equation}
Comparing the above with Equation \eqref{Eq:CNN}, given by
\begin{equation}
f(\X,\{ \W_i \}_{i=1}^2,\{ \b_i \}_{i=1}^2) = \ReLU \bigg( \ \W_2^T \ \ReLU \left( \ \W_1^T \X + \b_1 \ \right) + \b_2 \ \bigg),
\end{equation}
one can notice a striking similarity between the two. Moreover, by replacing $\P_\beta(\cdot)$ with the soft nonnegative thresholding, $\S_\beta^+(\cdot)$, we obtain that \textit{the aforementioned pursuit and the forward pass of the CNN are equal}! Notice that we are relying here on the discussion of Section \ref{Sec:Nonnegative}, where we have shown that the ReLU and the soft nonnegative thresholding are equal\footnote{A slight difference does exist between the soft nonnegative layered thresholding algorithm and the forward pass of the CNN. While in the former a constant threshold $\beta$ is employed for all entries, the latter uses a bias vector, $\b$, that might not be constant in all of its entries. This is of little significance, however, since a similar approach of an entry-based constant could be used in the layered thresholding algorithm as well.}.

Recall the optimization problem of the training stage of the CNN as shown in Equation \eqref{Eq:TrainingCNN}, given by
\begin{equation}
\underset{\{ \W_i \}_{i=1}^K,\{ \b_i \}_{i=1}^K,\U}{\min} \ \sum_j \ell \Big( h(\X_j), \U, f\left( \X_j,\{ \W_i \}_{i=1}^K,\{ \b_i \}_{i=1}^K \right) \Big),
\end{equation}
and its parallel in the ML-CSC model, the $\DLP$ problem, defined by
\begin{equation}
\underset{\{ \D_i \}_{i=1}^K,\U}{\min} \sum_j \ell \Big( h(\X_j),\U,\DCPstar(\X_j,\{ \D_i \}_{i=1}^K) \Big).
\end{equation}
Notice the remarkable similarity between both objectives, the only difference being in the feature vector on which the classification is done; in the CNN this is the output of the forward pass algorithm, given by $f\left( \X_j,\{ \W_i \}_{i=1}^K,\{ \b_i \}_{i=1}^K \right)$, while in the sparsity case this is the result of the $\DCP$ problem. In light of the discussion above, the solution for the $\DCP$ problem can be approximated using the layered thresholding algorithm, which is in turn equal to the forward pass of the CNN. We can therefore conclude that the problems solved by the training stage of the CNN and the $\DLP$ are tightly connected, and in fact are equal once the solution for the $\DLP$ is approximated via the layered thresholding algorithm (hence the name $\DLP$).

\section{Theoretical Study} \label{Sec:TheoreticalStudy}
Thus far, we have defined the ML-CSC model and its corresponding pursuits -- the $\DCP$ and $\DCPE$ problems. We have proposed a method to tackle them, coined the layered thresholding algorithm, which was shown to be equivalent to the forward pass of the CNN. Relying on this, we conclude that the proposed ML-CSC is the global Bayesian model implicitly imposed on the signal $\X$ when deploying the forward pass algorithm. Put differently, the ML-CSC answers the question of who are the signals belonging to the model behind the CNN. Having established the importance of our model, we now proceed to its theoretical analysis.

We should emphasize that the following study does not assume any specific form on the network's parameters, apart from a broad coherence property (as will be shown hereafter). This is in contrast to the work of \citep{bruna2013invariant} that assumes that the filters are Wavelets, or the analysis in \citep{giryes2015deep} that considers random weights.

\subsection[\texorpdfstring{Uniqueness of the $\DCP$ Problem}{Uniqueness of the DCP Problem}]{Uniqueness of the $\DCP$ Problem} \label{Sec:Uniqueness}

Consider a signal $\X$ admitting a multi-layer convolutional sparse representation defined by the sets $\{ \D_i \}_{i=1}^K$ and $\{\lambda_i\}_{i=1}^{K}$. Can another set of sparse vectors represent the signal $\X$? In other words, can we guarantee that, under some conditions, the set $\{\Gama_i\}_{i=1}^{K}$ is a unique solution to the $\DCP$ problem? In the following theorem we provide an answer to this question.

\begin{theorem}{(Uniqueness via the mutual coherence):} \label{Thm:Uniqueness}
	Consider a signal $\X$ satisfying the $\DCP$ model,
	\begin{align}
	\X & = \D_1\Gama_1 \\
	\Gama_1 & = \D_2\Gama_2 \\
	& \phantom{..} \vdots \\
	\Gama_{K-1} & = \D_K\Gama_K,
	\end{align}
	where $\{ \D_i \}_{i=1}^K$ is a set of convolutional dictionaries and $\left\{ \mu(\D_i) \right\}_{i=1}^K$ are their corresponding mutual coherences. If
	\begin{equation}
	\forall \ 1\leq i\leq K \qquad \| \Gama_i \|_{0,\infty}^\ss < \frac{1}{2} \left( 1 + \frac{1}{\mu(\D_i)} \right),
	\end{equation}
	then the set $\{\Gama_i\}_{i=1}^{K}$ is the unique solution to the $\DCP$ problem, assuming that the thresholds $\left\{ \lambda_i \right\}_{i=1}^K$ are chosen to satisfy
	\begin{equation}
	\forall \ 1\leq i\leq K \qquad \| \Gama_i \|_{0,\infty}^\ss \leq \lambda_i < \frac{1}{2} \left( 1 + \frac{1}{\mu(\D_i)} \right).
	\end{equation}
\end{theorem}

The proof for the above theorem is given in Appendix \ref{App:uniqueness}. In what follows, we present its importance in the context of CNN. Assume a signal $\X$ is fed into a network, resulting in a set of activation values across the different layers. These, in the realm of sparsity, correspond to the set of sparse representations $\{\Gama_i\}_{i=1}^{K}$, which according to the above theorem are in fact unique representations of the signal $\X$.


One might ponder at this point whether there exists an algorithm for obtaining the unique solution guaranteed in this subsection for the $\DCP$ problem. As previously mentioned, the layered thresholding algorithm is incapable of finding the exact representations, $\{\Gama_i\}_{i=1}^{K}$, due to the lack of a Least-Squares step after each layer. One should not despair, however, as we shall see in a following section an alternative algorithm, which manages to overcome this hurdle.

\subsection[\texorpdfstring{Global Stability of the $\DCPE$ Problem}{Global Stability of the DCP Problem}]{Global Stability of the $\DCPE$ Problem} \label{Sec:GlobalStability}
Consider an instance signal $\X$ belonging to the ML-CSC model, defined by the sets $\{ \D_i \}_{i=1}^K$ and $\{\lambda_i\}_{i=1}^{K}$. Assume $\X$ is contaminated by a noise vector $\E$, generating the perturbed signal $\Y=\X+\E$. Suppose we solve the $\DCPE$ problem and obtain a set of solutions $\{\hat{\Gama}_i\}_{i=1}^{K}$. How close is every solution in this set, $\hat{\Gama}_i$, to its corresponding true representation, $\Gama_i$? In what follows, we provide a theorem addressing this question of stability, the proof of which is deferred to Appendix \ref{App:GlobalStability}.

\begin{theorem}{(Stability of the solution to the $\DCPE$ problem):} \label{Thm:GlobalStability}
	Suppose a signal $\X$ that has a decomposition
	\begin{align}
	\X & = \D_1\Gama_1 \\
	\Gama_1 & = \D_2\Gama_2 \\
	& \phantom{..} \vdots \\
	\Gama_{K-1} & = \D_K\Gama_K
	\end{align}
	is contaminated with noise $\E$, where $\| \E \|_2 \leq \mathcal{E}_0$, resulting in $\Y = \X + \E$. For all $1\leq i\leq K$, if
	\begin{enumerate}
	\item $\| \Gama_i \|_{0,\infty}^\ss \leq \lambda_i < \frac{1}{2} \left( 1 + \frac{1}{\mu(\D_i)} \right)$; and
	\item ${\mathcal{E}_i}^2 = \frac{4\mathcal{E}_{i-1}^2}{1-(2\|\Gama_i\|_{0,\infty}^\ss-1)\mu(\D_i)}$,
	\end{enumerate}
	then 
	\begin{equation} \label{Eq:DCPEStability}
	\| \Gama_i- \hat{\Gama}_i \|_2^2 \leq {\mathcal{E}_i}^2,
	\end{equation}
	where the set $\{\hat{\Gama}_i\}_{i=1}^{K}$ is the solution for the $\DCPE$ problem.
\end{theorem}

Intuitively, the above claims that as long as all the feature vectors $\{\Gama_i\}_{i=1}^{K}$ are $\Loi$-sparse, then the representations obtained by solving the $\DCPE$ problem must be close to the true ones. Interestingly, the obtained bound increases as a function of the depth of the layer. This can be clearly seen from the recursive definition of $\mathcal{E}_i$, leading to the following bound
\begin{equation}
\|\Gama_i-\hat{\Gama}_i\|_2^2 \leq {\mathcal{E}_0}^2 \prod_{j=1}^{i} \frac{4}{1-(2\|\Gama_j\|_{0,\infty}^\ss-1)\mu(\D_j)}.
\end{equation}

Is this necessarily the true behavior of a deep network? Perhaps the answer to this resides in the choice we made above of considering the noise as adversary. A similar, yet somewhat more involved, analysis with a random noise assumption should be done, with the hope to see a better controlled noise propagation in this system. We leave this for our future work.

Another important remark is that the above bounds the \emph{absolute error} between the estimated and the true representation. In practice, however, the \emph{relative error} is of more importance. This is measured in terms of the signal to noise ratio (SNR), which we shall define in Section \ref{Sec:Experiments}.

Having established the stability of the $\DCPE$ problem, we now turn to the stability of the algorithms attempting to solve it, the chief one being the forward pass of CNN.

\subsection{Stability of the Layered Hard Thresholding} \label{Sec:StabilityLayeredHardThresholding}
Consider a signal $\X$ that admits a multi-layer convolutional sparse representation, which is defined by the sets $\{ \D_i \}_{i=1}^K$ and $\{\lambda_i\}_{i=1}^{K}$. Assume we run the layered hard thresholding algorithm on $\X$, obtaining the sparse vectors $\{\hat{\Gama}_i\}_{i=1}^{K}$. Under certain conditions, can we guarantee that the estimate $\hat{\Gama}_i$ recovers the true support of $\Gama_i$? or that the norm of the difference between the two is bounded? Assume $\X$ is contaminated with a noise vector $\E$, resulting in the measurement $\Y = \X + \E$. Assume further that this signal is then fed to the layered thresholding algorithm, resulting in another set of representations. How do the answers to the above questions change? To tackle these, we commence by presenting a stability claim for the simple hard thresholding algorithm, relying on the $\Loi$ norm. We should note that the analysis conducted in this subsection is for the noisy scenario, and the results for the noiseless case are simply obtained by setting the noise level to zero.

Next, we present a localized $\ell_2$ and $\ell_0$ measure of a global vector that will prove to be useful in the following analysis.
\begin{definition} \label{Def:LtinfLoip}
Define the $\| \cdot \|_{2,\infty}^\pp$ and $\| \cdot \|_{0,\infty}^\pp$ norm of $\Gama_i$ to be
\begin{equation}
\| \Gama_i \|_{2,\infty}^\pp = \max_j \| \PP_{i,j} \Gama_i \|_2
\end{equation}
and
\begin{equation}
\| \Gama_i \|_{0,\infty}^\pp = \max_j \| \PP_{i,j} \Gama_i \|_0,
\end{equation}
respectively.
The operator $\PP_{i,j}$ extracts the $j$-th patch of length $n_i m_i$ from the $i$-th sparse vector $\Gama_i$.
\end{definition}
In the above definition, the letter \textbf{p} emphasizes that the norms are computed by sweeping over all patches, rather than stripes. Recall that we have defined $m_0=1$, since the number of channels in the input signal $\X=\Gama_0$ is equal to one.

Given $\Y = \X + \E = \D_1\Gama_1 + \E$, the first stage of the layered hard thresholding algorithm attempts to recover the representation $\Gama_1$. Intuitively, assuming that the underlying representation $\Gama_1$ is $\Loi$-sparse, and that the energy of the noise $\E$ is $\Ltinf$-bounded; we would expect that the simple hard thresholding algorithm would succeed in recovering a solution $\hat{\Gama}_1$, which is both close to $\Gama_1$ and has its support. We now present such a claim, the proof of which is found in Appendix \ref{App:StabilityHardThresholding}.
\begin{lemma}{(Stable recovery of hard thresholding in the presence of noise):} \label{Thm:StabilityHardThresholding}
	Suppose a clean signal $\X$ has a convolutional sparse representation $\D_1\Gama_1$, and that it is contaminated with noise $\E$ to create the signal $\Y = \X + \E$, such that $\| \E \|_{2,\infty}^\pp \leq \epsilon_0$. Denote by $|\Gamma_1^{\text{min}}|$ and $|\Gamma_1^{\text{max}}|$ the lowest and highest entries in absolute value in $\Gama_1$, respectively. Denote further by $\hat{\Gama}_1$ the solution obtained by running the hard thresholding algorithm on $\Y$ with a constant $\beta_1$, i.e. $\hat{\Gama}_1 = \H_{\beta_1}(\D_1^T\Y)$. Assuming that
	\begin{enumerate} [\quad a) ]
	\item $\| \Gama_1 \|_{0,\infty}^\ss < \frac{1}{2} \left( 1 + \frac{1}{\mu(\D_1)} \frac{ |\Gamma_1^{\text{min}}| }{ |\Gamma_1^{\text{max}}| } \right) - \frac{1}{\mu(\D_1)}\frac{\epsilon_0}{|\Gamma_1^{\text{max}}|}$; and
	\item The threshold $\beta_1$ is chosen according to Equation \eqref{Eq:ConditionBeta1} (see below),
	\end{enumerate}
	then the following must hold:
	\begin{enumerate}
	\item The support of the solution $\hat{\Gama}_1$ is equal to that of $\Gama_1$; and
	\item $\| \Gama_1 - \hat{\Gama}_1 \|_{2,\infty}^\pp \leq \sqrt{\| \Gama_1 \|_{0,\infty}^\pp}\Big(\epsilon_0 + \mu(\D_1) \left( \| \Gama_1 \|_{0,\infty}^\ss - 1 \right) |\Gamma_1^{\text{max}}| \Big)$.
	\end{enumerate}
\end{lemma}

Notice that by plugging $\epsilon_0 = 0$ the above theorem covers the noiseless scenario. Notably, even in such a case, we obtain a deviation from the true representation due to the lack of a Least-Squares step.

We suspect that, both in the noiseless and the noisy case, the obtained bound might be improved, based on the following observation. Given an $\Ltinf$-norm bounded noise, the above proof first quantifies the deviation between the true representation and the estimated one in terms of the $\ell_\infty$ norm, and only then translates the latter into the $\Ltinf$ sense. A direct analysis going from an $\Ltinf$ input error to an $\Ltinf$ output deviation (bypassing the $\ell_\infty$ norm) might lead to smaller deviations. We leave this for future work. 

We now proceed to the next layer. Given $\hat{\Gama}_1$, which can be considered as a perturbed version of $\Gama_1$, the second stage of the layered hard thresholding algorithm attempts to recover the representation $\Gama_2$. Using the stability of the first layer -- guaranteeing that $\Gama_1$ and $\hat{\Gama}_1$ are close in terms of the $\Ltinf$ norm -- and relying on the $\Loi$-sparsity of $\Gama_2$, we show next that the second stage of the layered hard thresholding algorithm is stable as well. Applying the same rationale to all the remaining layers, we obtain the theorem below guaranteeing the stability of the complete layered hard thresholding algorithm.

\begin{theorem}{(Stability of layered hard thresholding in the presence of noise):} \label{Thm:StabilityLayeredHardThresholding}
	Suppose a clean signal $\X$ has a decomposition 
	\begin{align}
	\X & = \D_1\Gama_1 \\
	\Gama_1 & = \D_2\Gama_2 \\
	& \phantom{..} \vdots \\
	\Gama_{K-1} & = \D_K\Gama_K,
	\end{align}
	and that it is contaminated with noise $\E$ to create the signal $\Y = \X + \E$, such that $\| \E \|_{2,\infty}^\pp \leq \epsilon_0$. Denote by $|\Gamma_i^{\text{min}}|$ and $|\Gamma_i^{\text{max}}|$ the lowest and highest entries in absolute value in the vector $\Gama_i$, respectively. Let $\{\hat{\Gama}_i\}_{i=1}^{K}$ be the set of solutions obtained by running the layered hard thresholding algorithm with thresholds $\{\beta_i\}_{i=1}^{K}$, i.e. $\hat{\Gama}_i=\H_{\beta_i}(\D_i^T\hat{\Gama}_{i-1})$ where $\hat{\Gama}_{0}=\Y$.
	 Assuming that $\forall \ 1 \leq i \leq K$
	 \begin{enumerate} [\quad a) ]
	 	\item $\| \Gama_i \|_{0,\infty}^\ss < \frac{1}{2} \left( 1 + \frac{1}{\mu(\D_i)} \frac{ |\Gamma_i^{\text{min}}| }{ |\Gamma_i^{\text{max}}| } \right) - \frac{1}{\mu(\D_i)}\frac{ \epsilon_{i-1} }{|\Gamma_i^{\text{max}}|}$; and
	 	\item The threshold $\beta_i$ is chosen according to Equation \eqref{Eq:ConditionBeta_i},
	 \end{enumerate}
	 then\footnote{Recall that $\| \Gama_i \|_{2,\infty}^\pp$ is defined to be the maximal $\ell_2$ norm of a patch extract from $\Gama_i$. The size of this patch is defined according to the dictionary $\D_{i+1}$. However, the last sparse vector $\Gama_K$ does not have a corresponding dictionary $\D_{K+1}$. As such, the size of a patch in $\Gama_K$ can be chosen arbitrarily. Where the choice of the size directly affects the bound on the difference, $\epsilon_i$, due to the term $\sqrt{\| \Gama_i \|_{0,\infty}^\pp}$.}
	 \begin{enumerate}
	 	\item The support of the solution $\hat{\Gama}_i$ is equal to that of $\Gama_i$; and
	 	\item $\| \Gama_i - \hat{\Gama}_i \|_{2,\infty}^\pp \leq \epsilon_i$,
	 \end{enumerate}
	where $\epsilon_i = \sqrt{ \| \Gama_{i} \|_{0,\infty}^\pp } \ \Big( \epsilon_{i-1} + \mu(\D_i) \left( \| \Gama_i \|_{0,\infty}^\ss - 1 \right) |\Gamma_i^{\text{max}}| \Big)$.
\end{theorem}

The proof for the above is given in Appendix \ref{App:StabilityLayeredHardThresholding}. We now turn to an analogous theorem for the forward pass of the CNN, prior to discussing the surprising implications of these theorems.

\subsection{Stability of the Forward Pass (Layered Soft Thresholding)}
In light of the discussion in Section \ref{Sec:Crux}, the equivalence between the layered thresholding algorithm and the forward pass of the CNN is achieved assuming that the operator employed is the nonnegative soft thresholding $\S_\beta^+(\cdot)$. However, thus far, we have analyzed the closely related hard version $\H_\beta(\cdot)$ instead. In what follows, we show how the stability theorem presented in the previous subsection can be modified to the soft version, $\S_\beta(\cdot)$. For simplicity, and in order to stay in line with the vast sparse representation theory, herein we choose not to assume the nonnegative assumption. This implies that we are proposing a slightly different CNN architecture in which the ReLU function is two sided \citep{kavukcuoglu2010learning}. We now move to the stable recovery of the soft thresholding algorithm.

\begin{lemma}{(Stable recovery of soft thresholding in the presence of noise):} \label{Thm:StabilitySoftThresholding}
	Suppose a clean signal $\X$ has a convolutional sparse representation $\D_1\Gama_1$, and that it is contaminated with noise $\E$ to create the signal $\Y = \X + \E$, such that $\| \E \|_{2,\infty}^\pp \leq \epsilon_0$. Denote by $|\Gamma_1^{\text{min}}|$ and $|\Gamma_1^{\text{max}}|$ the lowest and highest entries in absolute value in $\Gama_1$, respectively. Denote further by $\hat{\Gama}_1$ the solution obtained by running the soft thresholding algorithm on $\Y$ with a constant $\beta_1$, i.e. $\hat{\Gama}_1 = \S_{\beta_1}(\D_1^T\Y)$. Assuming that
	\begin{enumerate} [\quad a) ]
	\item $\| \Gama_1 \|_{0,\infty}^\ss < \frac{1}{2} \left( 1 + \frac{1}{\mu(\D_1)} \frac{ |\Gamma_1^{\text{min}}| }{ |\Gamma_1^{\text{max}}| } \right) - \frac{1}{\mu(\D_1)}\frac{\epsilon_0}{|\Gamma_1^{\text{max}}|}$; and \label{Assumption:SoftThreshAssumption1}
	\item The threshold $\beta_1$ is chosen according to Equation \eqref{Eq:ConditionBeta1}, \label{Assumption:SoftThreshAssumption2}
	\end{enumerate}
	then the following must hold:
	\begin{enumerate}
	\item The support of the solution $\hat{\Gama}_1$ is equal to that of $\Gama_1$; and
	\item $\| \Gama_1 - \hat{\Gama}_1 \|_{2,\infty}^\pp \leq \sqrt{\| \Gama_1 \|_{0,\infty}^\pp}\Big(\epsilon_0 + \mu(\D_1) \left( \| \Gama_1 \|_{0,\infty}^\ss - 1 \right) |\Gamma_1^{\text{max}}| + \beta_1 \Big)$.
	\end{enumerate}
\end{lemma}

Armed with the above lemma, which is proven in Appendix \ref{App:StabilitySoftThresholding}, we now proceed to the stability of the forward pass of the CNN.

\begin{theorem}{(Stability of the forward pass (layered soft thresholding algorithm) in the presence of noise):} \label{Thm:StabilityLayeredSoftThresholding}
	Suppose a clean signal $\X$ has a decomposition 
	\begin{align}
	\X & = \D_1\Gama_1 \\
	\Gama_1 & = \D_2\Gama_2 \\
	& \phantom{..} \vdots \\
	\Gama_{K-1} & = \D_K\Gama_K,
	\end{align}
	and that it is contaminated with noise $\E$ to create the signal $\Y = \X + \E$, such that $\| \E \|_{2,\infty}^\pp \leq \epsilon_0$. Denote by $|\Gamma_i^{\text{min}}|$ and $|\Gamma_i^{\text{max}}|$ the lowest and highest entries in absolute value in the vector $\Gama_i$, respectively. Let $\{\hat{\Gama}_i\}_{i=1}^{K}$ be the set of solutions obtained by running the layered soft thresholding algorithm with thresholds $\{\beta_i\}_{i=1}^{K}$, i.e. $\hat{\Gama}_i=\S_{\beta_i}(\D_i^T\hat{\Gama}_{i-1})$ where $\hat{\Gama}_{0}=\Y$.
	Assuming that $\forall \ 1 \leq i \leq K$
	\begin{enumerate} [\quad a) ]
		\item $\| \Gama_i \|_{0,\infty}^\ss < \frac{1}{2} \left( 1 + \frac{1}{\mu(\D_i)} \frac{ |\Gamma_i^{\text{min}}| }{ |\Gamma_i^{\text{max}}| } \right) - \frac{1}{\mu(\D_i)}\frac{ \epsilon_{i-1} }{|\Gamma_i^{\text{max}}|}$; and
		\item The threshold $\beta_i$ is chosen according to Equation \eqref{Eq:ConditionBeta_i} (with the $\epsilon_i$ defined below),
	\end{enumerate}
	then
	\begin{enumerate}
		\item The support of the solution $\hat{\Gama}_i$ is equal to that of $\Gama_i$; and
		\item $\| \Gama_i - \hat{\Gama}_i \|_{2,\infty}^\pp \leq \epsilon_i$,
	\end{enumerate}
	where $\epsilon_i = \sqrt{ \| \Gama_{i} \|_{0,\infty}^\pp } \ \Big( \epsilon_{i-1} + \mu(\D_i) \left( \| \Gama_i \|_{0,\infty}^\ss - 1 \right) |\Gamma_i^{\text{max}}| + \beta_i \Big)$.
\end{theorem}

The proof for the above is omitted since it is tantamount to that of Theorem \ref{Thm:StabilityLayeredHardThresholding}. As one can see, the layered soft thresholding algorithm is in fact inferior to its hard variant due to the added constant of $\beta_i$ in the local error level, $\epsilon_i$. This results in a more strict assumption on the $\Loi$ norm of the various representations and also augments the bound on the distance between the true sparse vector and the one recovered. Following this observation, a natural question arises; why does the deep learning community employ the ReLU, which corresponds to a soft nonnegative thresholding operator instead of another nonlinearity that is more similar to its hard counterpart? One possible explanation could be that the filter training stage of the CNN becomes harder when the ReLU is replaced with a non-convex alternative, which also has discontinuities, such as the hard thresholding operator.

The above theorem guarantees that the distances between the original representations and the ones obtained from the CNN are bounded. Even if we set $\epsilon_0 = 0$, the recovered activations deviate from the true ones, simply because the layered thresholding algorithm does not do a perfect job, even on a noiseless signal. When the signal is noisy, these deviations are strengthened, but still in a controlled way.

This, by itself, might not be surprising. After all, the CNN is a deterministic system of linear operations (convolutions), followed by simple non-linearities that are non-expanding. If we feed a slightly perturbed signal to such a system, it is clear that the activations all along the network will be perturbed as well with a bounded effect. However, the above theorem shows far more than that. There are, in fact, two types of stabilities, the trivial one that considers the sensitivity of the whole feed-forward network to perturbations in its input, and the more intricate one that shows that this system enables a rather accurate recovery of the \textbf{generating representations}. The second option is the stability we prove here.

\subsection{Guarantees for Fully Connected Networks}

One should note that the convolutional structure imposed on the dictionaries in our model could be removed, and the theoretical guarantees we have provided above would still hold. The reason being is that the unconstrained dictionary can be regarded as a convolutional one, constructed from a single shift of a local matrix with no circular boundary. In the context of CNN, this is analogous to a fully connected layer. As such, the theoretical analysis provided here sheds light on both convolutional and fully connected networks. A different point of view on the same matter can also be proposed; fully connected layers can be viewed as convolutional ones with filters that cover their entire input \citep{long2015fully}.

\section{Layered Basis Pursuit -- The Future of Deep Learning?} \label{Sec:Future}
The stability analysis presented above unveils two significant limitations of the forward pass of the CNN. First, this algorithm is incapable of recovering the unique solution for the $\DCP$ problem, the existence of which is guaranteed from Theorem \ref{Thm:Uniqueness}. This acts against our expectations, since in the traditional sparsity inspired model it is a well known fact that such a unique representation can be retrieved, assuming certain conditions are met.

The second issue is with the condition for the successful recovery of the true support. The $\Loi$ norm of the true solution, $\Gama_i$, is required to be less than an expression that depends on the term $|\Gamma_i^{\text{min}}|/|\Gamma_i^{\text{max}}|$. The dependence on this ratio is a direct consequence of the forward pass algorithm relying on the simple thresholding operator that is known for having such a theoretical limitation\footnote{The dependence on the ratio is also a direct consequence of assuming a worst-case analysis. Perhaps in reality this ratio does not play such a critical role.}. However, alternative pursuits whose success would not depend on this ratio could be proposed, as indeed was done in the Sparse-Land model; resulting in both theoretical and practical benefits.

A solution for the first problem, already presented throughout this work, is a two-stage approach. First, run the thresholding operator in order to recover the correct support. Then, once the atoms are chosen, their corresponding coefficients can be obtained by solving a linear system of equations. In addition to retrieving the true representation in the noiseless case, this step can also be beneficial in the noisy scenario, resulting in a solution closer to the underlying one. However, since no such step exists in current CNN architectures, we refrain from further analyzing its theoretical implications.

Next, we present an alternative to the layered soft thresholding algorithm, which will tackle both of the aforementioned problems. Recall that the result of the soft thresholding is a simple approximation of the solution for the $\Pone$ problem, previously defined in Equation \eqref{Eq:P1}. In every layer, instead of applying a simple thresholding operator that estimates the sparse vector by computing $\hat{\Gama}_i = \S_{\beta_i}(\D_i^T\hat{\Gama}_{i-1})$; we propose to tackle the full pursuit, i.e. to minimize
\begin{equation} \label{Eq:LayeredBP}
\hat{\Gama}_{i} = \underset{\Gama_i}{\arg\min} \ \|\Gama_i\|_1 \ \text{ s.t. } \  \hat{\Gama}_{i-1} = \D_i\Gama_i.
\end{equation}
Notice that one could readily obtain the nonnegative sparse coding problem by simply adding an extra constraint in the above equation, forcing the coefficients in $\Gama_i$ to be nonnegative. More generally, Equation \eqref{Eq:LayeredBP} can be written in its Lagrangian formulation
\begin{equation} \label{Eq:LagrangianBP}
\hat{\Gama}_{i} = \underset{\Gama_i}{\arg\min} \ \xi_i \|\Gama_i\|_1 + \frac{1}{2}\| \D_i\Gama_i - \hat{\Gama}_{i-1} \|_2^2,
\end{equation}
where the constant $\xi_i$ is proportional to the noise level and should tend to zero in the noiseless scenario. We name the above the \emph{layered basis pursuit} (BP) algorithm. In practice, one possible method for solving it is the iterative soft thresholding (IST). Formally, this obtains the minimizer of Equation \eqref{Eq:LagrangianBP} by repeating the following recursive formula
\begin{equation} \label{Eq:IT}
\hat{\Gama}_i^t = \S_{\xi_i/c_i}\left( \hat{\Gama}_i^{t-1} + \frac{1}{c_i}\D_i^T \left( \hat{\Gama}_{i-1} - \D_i \hat{\Gama}_i^{t-1} \right) \right),
\end{equation}
where $\hat{\Gama}_i^t$ is the estimate of $\Gama_i$ at iteration $t$. The above can be interpreted as a simple projected gradient descent algorithm, where the constant $c_i$ is inversely proportional to its step size. As a result, if $c_i$ is chosen to be large enough\footnote{The constant $c_i$ should satisfy $c_i > 0.5\lambda_\text{max} \left(\D_i^T \D_i \right)$, where $\lambda_\text{max} \left(\D_i^T \D_i\right)$ is the maximal eigenvalue of the gram matrix $\D_i^T \D_i$ \citep{combettes2005signal}.}, the above algorithm is guaranteed to converge to its global minimum that is the solution of \eqref{Eq:LagrangianBP}, as was shown in \citep{Daubechies2004}. The method obtained by gradually computing the set of sparse representations, $\{\Gama_i\}_{i=1}^{K}$, via the IST is summarized in Algorithm \ref{Alg:LayeredIT} and named \emph{layered iterative soft thresholding}. Notice that this algorithm coincides with the simple layered soft thresholding if it is run for a single iteration with $c_i=1$ and initialized with $\hat{\Gama}_i^0 = \mathbf{0}$. This implies that the above algorithm is a natural extension to the forward pass of the CNN.

With respect to the computational aspects of the IST algorithm, the work of \citep{gregor2010learning} proposed the LISTA method, showing how the number of iterations required by the IST to convergence can be reduced using neural networks. Analogously, the work of \citep{xin2016maximal} presented a generalization of the iterative hard thresholding (IHT), which was shown both theoretically and empirically to be superior to the original IHT.

\begin{algorithm}[t]
	\caption{The layered iterative soft thresholding algorithm.}
	\label{Alg:LayeredIT}
	
	\textbf{Input:}
	\begin{algorithmic}
	\State $\X$ -- a signal.
	\State $\{ \D_i \}_{i=1}^K$ -- convolutional dictionaries.
	\State $\P \in \{ \S, \S^+ \}$ -- a soft thresholding operator.
	\State $\{\xi_i\}_{i=1}^{K}$ -- Lagrangian parameters.
	\State $\{1/c_i\}_{i=1}^{K}$ -- step sizes.
	\State $\{T_i\}_{i=1}^{K}$ -- number of iterations.
	\end{algorithmic}
	
	\textbf{Output:}
	\begin{algorithmic}
	\State A set of representations $\{\hat{\Gama}_i\}_{i=1}^{K}$.
	\end{algorithmic}
	
	\textbf{Process:}
	\begin{algorithmic}[1]
	\State $\hat{\Gama}_0 \leftarrow \X$
	\For{$i=1:K$}
		\State $\hat{\Gama}_i^0 \leftarrow \mathbf{0}$
		\For{$t=1:T_i$}
			\State $\hat{\Gama}_i^t \leftarrow \P_{\xi_i/c_i}\left( \hat{\Gama}_i^{t-1} + \frac{1}{c_i}\D_i^T \left( \hat{\Gama}_{i-1} - \D_i \hat{\Gama}_i^{t-1} \right) \right)$
		\EndFor
		\State $\hat{\Gama}_i \leftarrow \hat{\Gama}_i^{T_i}$
	\EndFor
	\end{algorithmic}
	
\end{algorithm}

The original motivation for the layered IST was its theoretical superiority over the forward pass algorithm -- one that will be explored in detail in the next subsection. Yet more can be said about this algorithm and the CNN architecture it induces. In \citep{gregor2010learning} it was shown that the IST algorithm can be formulated as a simple recurrent neural network. As such, the same can be said regarding the layered IST algorithm proposed here, with the exception that the induced recurrent network is much deeper. The reader can therefore interpret this part of the work as a theoretical study of a special case of recurrent neural networks.

From another perspective, the underlying architecture of the layered IST algorithm is a cascade of $K$ blocks. Each of these corresponds to a fixed number of unfolded iterations, $T_i$, of a single IST algorithm. As such, it contains several convolutional layers with shared weights, as well as skip connections in order to compute the residual, $\hat{\Gama}_{i-1} - \D_i \hat{\Gama}_i^{t-1}$, as defined in Equation \eqref{Eq:IT}. Interestingly, the above description is reminiscent (though not exact) of residual networks \citep{he2015deep}, which have recently led to state-of-the-art results in image recognition.

\subsection{Success of Layered BP Algorithm}
In Section \ref{Sec:Uniqueness}, we established the uniqueness of the solution for the $\DCP$ problem, assuming that certain conditions on the $\Loi$ norm of the underlying representations are met. However, as we have seen in the theoretical analysis of the previous section, the forward pass of the CNN is incapable of finding this unique solution; instead, it is guaranteed to be close to it in terms of the $\Ltinf$ norm. Herein, we address the question of whether the layered BP algorithm can prevail in a task where the forward pass did not.

\begin{theorem}{(Layered BP recovery guarantee using the $\Loi$ norm):} \label{Thm:SuccessLayeredBP}
	Consider a signal $\X$,
	\begin{align}
	\X & = \D_1\Gama_1 \\
	\Gama_1 & = \D_2\Gama_2 \\
	& \phantom{..} \vdots \\
	\Gama_{K-1} & = \D_K\Gama_K,
	\end{align}
	where $\{ \D_i \}_{i=1}^K$ is a set of convolutional dictionaries and $\left\{ \mu(\D_i) \right\}_{i=1}^K$ are their corresponding mutual coherences. Assuming that $\forall \ 1\leq i\leq K$
	\begin{equation}
	\| \Gama_i \|_{0,\infty}^\ss < \frac{1}{2} \left( 1 + \frac{1}{\mu(\D_i)} \right)
	\end{equation}
	then the layered BP algorithm is guaranteed to recover the set $\{\Gama_i\}_{i=1}^{K}$.
\end{theorem}

The proof for the above can be directly derived from the recovery condition of the BP using the $\Loi$ norm, as presented in \citep{Papyan2016_1}. The implications of this theorem are that the layered BP algorithm can indeed recover the unique solution to the $\DCP$ problem.

\subsection{Stability of Layered BP Algorithm}
Having established the guarantee for the success of the layered BP algorithm, we now move to its stability analysis. In particular, in a noisy scenario where obtaining the true underlying representations is impossible, does this algorithm remain stable? If so, how do its guarantees compare to those of the layered thresholding algorithm? The following theorem, which we prove in Appendix \ref{App:StabilityLayeredBP}, aims to answer these questions.

\begin{theorem}{(Stability of the layered BP algorithm in the presence of noise):} \label{Thm:StabilityLayeredBP}
	Suppose a clean signal $\X$ has a decomposition 
	\begin{align}
	\X & = \D_1\Gama_1 \\
	\Gama_1 & = \D_2\Gama_2 \\
	& \phantom{..} \vdots \\
	\Gama_{K-1} & = \D_K\Gama_K,
	\end{align}
	and that it is contaminated with noise $\E$ to create the signal $\Y = \X + \E$, such that $\| \E \|_{2,\infty}^\pp \leq \epsilon_0$. Let $\{\hat{\Gama}_i\}_{i=1}^{K}$ be the set of solutions obtained by running the layered BP algorithm with parameters $\{\xi_i\}_{i=1}^{K}$. Assuming that $\ \forall \  1\leq i\leq K$
	\begin{enumerate} [\quad a) ]
	\item $\| \Gama_i \|_{0,\infty}^\ss < \frac{1}{3} \left( 1 + \frac{1}{\mu(\D_i)} \right)$; and
	\item $\xi_i = 4 \epsilon_{i-1}$,
	\end{enumerate}
	then
	\begin{enumerate}
	\item The support of the solution $\hat{\Gama}_i$ is contained in that of $\Gama_i$;
	\item $\| \Gama_i - \hat{\Gama}_i \|_{2,\infty}^\pp \leq \epsilon_i$;
	\item In particular, every entry of $\Gama_i$ greater in absolute value than $\frac{\epsilon_i}{\sqrt{ \| \Gama_{i} \|_{0,\infty}^\pp }}$ is guaranteed to be recovered; and
	\item The solution $\hat{\Gama}_i$ is the unique minimizer of the Lagrangian BP problem (Equation \eqref{Eq:LagrangianBP}),
	\end{enumerate}
	where $\epsilon_i = \| \E \|_{2,\infty}^\pp \ 7.5^i \ \prod_{j=1}^i\sqrt{ \| \Gama_{j} \|_{0,\infty}^\pp }$.
\end{theorem}

Several remarks are due at this point. The condition for the stability of the layered thresholding algorithm, given by
\begin{equation}
\| \Gama_i \|_{0,\infty}^\ss < \frac{1}{2} \left( 1 + \frac{1}{\mu(\D_i)} \frac{ |\Gamma_i^{\text{min}}| }{ |\Gamma_i^{\text{max}}| } \right) - \frac{1}{\mu(\D_i)}\frac{ \epsilon_{i-1} }{|\Gamma_i^{\text{max}}|},
\end{equation}
is expected to be more strict than that of the theorem presented above, which is
\begin{equation}
\| \Gama_i \|_{0,\infty}^\ss < \frac{1}{3} \left( 1 + \frac{1}{\mu(\D_i)} \right).
\end{equation}
In the case of the layered BP algorithm, the bound on the $\Loi$ norm of the underlying sparse vectors does no longer depend on the ratio $|\Gamma_i^{\text{min}}|/|\Gamma_i^{\text{max}}|$ -- a term present in all the theoretical results of the thresholding algorithm. Moreover, the $\Loi$ norm becomes independent of the local noise level of the previous layers, thus allowing more non-zeros per stripe.

In addition, similar to the stability analysis presented in Section \ref{Sec:GlobalStability}, the above shows the growth (as a function of the depth) of the distance between the recovered representations and the true ones.

\section{A Closer Look at the Proposed Model} \label{Sec:CloserLook}
In this section, we revisit the assumptions of our model by imposing additional constraints on the dictionaries involved and showing their theoretical benefits. These additional assumptions originate from the current common practice of both CNN and sparsity.

\subsection{When a Patch Becomes a Stripe} \label{Sec:SparseDictionary}
Throughout the analysis presented in this work, we have assumed that the representations in the different layers, $\{\Gama_i\}_{i=1}^{K}$, are $\Loi$-sparse. Herein, we study the propagation of the $\Loi$ norm throughout the layers of the network, showing how an assumption on the sparsity of the deepest representation $\Gama_K$ reflects on that of the remaining layers. The exact connection between the sparsities will be given in terms of a simple characterization of the dictionaries $\{ \D_i \}_{i=1}^K$.

Consider the representation $\Gama_{K-1}$, given by
\begin{equation}
\Gama_{K-1} = \D_K \Gama_K,
\end{equation}
where $\Gama_K$ is $\Loi$-sparse. Following Figure \ref{Fig:LocalSystem}, the $i$-th patch in $\Gama_{K-1}$ can be expressed as
\begin{equation}
\PP_{K-1,i} \Gama_{K-1} = \O_K \hspace{0.05cm} \gama_{K,i},
\end{equation}
where $\O_K$ is the stripe-dictionary of $\D_K$, the vector $\PP_{K-1,i}\Gama_{K-1}$ is the $i$-th patch in $\Gama_{K-1}$ and $\gama_{K,i}$ is its corresponding stripe. Recalling the definition of the $\| \cdot \|_{0,\infty}^\pp$ norm (Definition \ref{Def:LtinfLoip} in Section \ref{Sec:StabilityLayeredHardThresholding}), we have that
\begin{equation}
\| \Gama_{K-1} \|_{0,\infty}^\pp = \max_i \| \O_K \hspace{0.05cm} \gama_{K,i} \|_0.
\end{equation}
Consider the following definition.
\begin{definition}
Define the induced $\ell_0$ pseudo-norm of a dictionary $\D$, denoted by $\| \D \|_0$, to be the maximal number of non-zeros in any of its atoms\footnote{According to the definition of the induced norm $\| \D \|_0 = \max_\v \| \D\v \|_0 \ \ \text{s.t.} \ \| \v \|_0 = 1$. Since $\| \v \|_0 = 1$, the multiplication $\D\v$ is simply equal to one of the atoms in $\D$ times a scalar, and $\| \D\v \|_0$ counts the number of non-zeros in this atom. As a result, $\| \D \|_0$ is equal to the maximal number of non-zeros in any atom from $\D$.}.
\end{definition}
\noindent
The multiplication $\O_K \hspace{0.05cm} \gama_{K,i}$ can be seen as a linear combination of at most $\| \gama_{K,i} \|_0$ atoms, each contributing no more than $\| \O_K \|_0$ non-zeros. As such
\begin{equation}
\| \Gama_{K-1} \|_{0,\infty}^\pp \leq \max_i \| \O_K \|_0 \ \| \gama_{K,i} \|_0.
\end{equation}
Noticing that $\| \O_K \|_0 = \| \D_K \|_0$ (as can be seen in Figure \ref{Fig:LocalSystem}), and using the definition of the $\| \cdot \|_{0,\infty}^\ss$ norm, we conclude that
\begin{equation} \label{Eq:RelationLoi}
\| \Gama_{K-1} \|_{0,\infty}^\pp \leq \| \D_K \|_0 \ \| \Gama_K \|_{0,\infty}^\ss.
\end{equation}
In other words, given $\| \Gama_K \|_{0,\infty}^\ss$ and $\| \D_K \|_0$, we can bound the maximal number of non-zeros in a patch from $\Gama_{K-1}$.

The claims in Section \ref{Sec:TheoreticalStudy} and \ref{Sec:Future} are given in terms of not only $\| \Gama_{K-1} \|_{0,\infty}^\pp$, but also $\| \Gama_{K-1} \|_{0,\infty}^\ss$. According to Table \ref{Table:Notations}, the length of a patch in $\Gama_{K-1}$ is $n_{K-1}m_{K-1}$, while the size of a stripe is $(2n_{K-2}-1)m_{K-1}$. As such, we can fit $(2n_{K-2}-1) / n_{K-1}$ patches in a stripe. Assume for simplicity that this ratio is equal to one. As a result, we obtain that a patch in the \emph{signal} $\Gama_{K-1}$ extracted from the system
\begin{equation}
\Gama_{K-1} = \D_K\Gama_K,
\end{equation}
is also a stripe in the \emph{representation} $\Gama_{K-1}$ when considering
\begin{equation}
\Gama_{K-2} = \D_{K-1}\Gama_{K-1},
\end{equation}
hence the name of this subsection. Leveraging this assumption, we return to Equation \eqref{Eq:RelationLoi} and obtain that
\begin{equation}
\| \Gama_{K-1} \|_{0,\infty}^\ss = \| \Gama_{K-1} \|_{0,\infty}^\pp \leq \| \D_K \|_0 \ \| \Gama_K \|_{0,\infty}^\ss.
\end{equation}
Using the same rationale for the remaining layers, and assuming that once again the patches become stripes, we conclude that
\begin{equation} \label{Eq:LoiPropagation}
\| \Gama_i \|_{0,\infty}^\ss = \| \Gama_i \|_{0,\infty}^\pp \leq \| \Gama_K \|_{0,\infty}^\ss \prod_{j=i+1}^{K} \| \D_j \|_0.
\end{equation}
We note that our assumption here of having sparse dictionaries is reasonable, since at the training stage of the CNN an $\ell_1$ penalty is often imposed on the filters as a regularization, promoting their sparsity. The conclusion thus is that the $\Loi$ norm is expected to decrease as a function of the depth of the representation. This aligns with the intuition that the higher the depth, the more abstraction one obtains in the filters, and thus the less non-zeros are required to represent the data. Taking this to the extreme, if every input signal could be represented via a single coefficient at the deepest layer, we would obtain that its $\Loi$ norm is equal to one.

\subsection{On the Role of the Spatial-Stride} \label{Sec:Stride}
A common step among practitioners of CNN \citep{krizhevsky2012imagenet,simonyan2014very,he2015deep} is to convolve the input to each layer with a set of filters, skipping a fixed number of spatial locations in a regular pattern. One of the primary motivations for this is to reduce the dimensions of the kernel maps throughout the layers, leading to \textbf{computational benefits}. In this subsection we unveil some \textbf{theoretical benefits} of this common practice, which we coin \emph{spatial-stride}.

Following Figure \ref{Fig:GlobalSystem}, recall that $\D_i$ is a stride convolutional dictionary that skips $m_{i-1}$ shifts at a time, which correspond to the number of channels in $\Gama_{i-1}$. Translating the spatial-stride to our language, the above mentioned works do not consider all spatial shifts of the filters in $\D_i$. Instead, a stride of $m_{i-1}s_{i-1}$ is employed, where $m_{i-1}$ corresponds to the \emph{channel-stride}, while $s_{i-1}$ is due to the \emph{spatial-stride}. The addition of the latter implies that instead of assuming that the $i$-th sparse vector satisfies $\Gama_{i-1}=\D_i\Gama_i$, we have that $\Q_{i-1}\Gama_{i-1}=\D_i\Q_i^T\Q_i\Gama_i$. We denote $\Q_i^T\in\mathbb{R}^{Nm_i \times Nm_i/s_{i-1}}$ as a columns' selection operator that chooses the atoms from $\D_i$ that align with the spatial-stride. The coefficients corresponding to these atoms are extracted from $\Gama_i$ (resulting in its subsampled version) via the $\Q_i$ matrix. In light of the above discussion, we modify the $\DCP$ problem, as defined in Definition \ref{Def:DCP}, into the following
\begin{align*}
\hspace{1.5cm}\text{find} \quad \{\Gama_i\}_{i=1}^{K} \qquad \text{ s.t. } \qquad
\X					& = \D_1\Q_1^T \Q_1\Gama_1,		& \| \Q_1\Gama_1 \|_{0,\infty}^\ss		& \leq \lambda_1 \\
\Q_1\Gama_1			& = \D_2\Q_2^T \Q_2\Gama_2,		& \| \Q_2\Gama_2 \|_{0,\infty}^\ss		& \leq \lambda_2 \\
             		& \phantom{..} \vdots			&										& \phantom{..} \vdots \\
\Q_{K-1}\Gama_{K-1}	& = \D_K\Q_K^T \Q_K\Gama_K,		& \| \Q_K\Gama_{K} \|_{0,\infty}^\ss	& \leq \lambda_K.\hspace{1.5cm}
\end{align*}
Note that while the original $\| \Gama_i \|_{0,\infty}^\ss$ is equal to the maximal number of non-zeros in a stripe of length \mbox{$(2 n_{i-1} - 1)m_i$} in $\Gama_i$, the term $\| \Q_i\Gama_i \|_{0,\infty}^\ss$ counts the same quantity but for stripes of length \mbox{$(2 \ \ceil*{n_{i-1}/s_{i-1}} - 1)m_i$} in $\Q_i\Gama_i$.

According to the study in Section \ref{Sec:TheoreticalStudy} and \ref{Sec:Future}, the theoretical advantage of the spatial-stride is twofold. First, consider the mutual coherence of the stride convolutional dictionary $\D_i$. Due to the locality of the filters and their restriction to certain spatial shifts, the mutual coherence of $\D_i\Q_i^T$ is expected to be lower than that of $\D_i$, thus leading to more non-zeros allowed per stripe. Second, the length of a stripe in $\Q_i\Gama_i$ is equal to $(2 \ \ceil*{n_{i-1}/s_{i-1}} - 1)m_i$, while that of $\Gama_i$ is $(2 n_{i-1} - 1)m_i$. As such, our analysis allows a larger number of non-zeros per a smaller-sized stripe. From another perspective, notice that imposing a spatial-stride on the dictionary $\D_i$ is equivalent to forcing a portion of the entries in $\Gama_i$ to be zero. As such, the spatial-stride encourages sparser solutions.

\section{Experiments: The Generator Behind the CNN} \label{Sec:Experiments}
Consider the following question: can we synthesize signals obeying the ML-CSC model? Throughout this work we have posited that the answer to this question is positive; we have assumed the existence of a set of signals $\X$, which satisfy $\forall i \ \Gama_i=\D_{i+1}\Gama_{i+1}$ where $\{\Gama_i\}_{i=1}^K$ are all $\Loi$ bounded. However, a natural question arises as to whether we can give a simple example of set of dictionaries $\{\D_i\}_{i=1}^K$ and their corresponding signals $\X$ that indeed satisfy our model assumptions.

A na\"ive attempt would be to choose an arbitrary set of dictionaries, $\left\{ \D_i \right\}_{i=1}^K$, and a random deepest representation, $\Gama_K$, and compute the remaining sparse vectors (and the signal itself) using the set of relations $\Gama_i=\D_{i+1}\Gama_{i+1}$. However, without further restrictions, this would lead to a set of representations $\left\{ \Gama_i \right\}_{i=1}^K$ with growing $\Loi$ norm as we propagate towards $\Gama_0$. A somewhat better approach would be to impose sparsity on the dictionaries involved, as suggested in Section \ref{Sec:SparseDictionary}, thus leading to sparser representations. However, besides the obvious drawback of forcing a limiting structure on the dictionaries, as can be seen in Equation \eqref{Eq:LoiPropagation}, in the worst case this would also lead to growth in the density of the representations, even if it is more controlled. The spatial-stride -- at first glance unrelated to this discussion -- is another solution that addresses the same problem. In particular, in Section \ref{Sec:Stride} this idea was shown to encourage sparser vectors by forcing zeros in a regular pattern in the set of representations $\{\Gama_i\}_{i=1}^K$.

In this section we combine the above notions in order to achieve our goal -- generate a set of signals that will satisfy the ML-CSC assumptions. These will then serve as a playground for several experiments, which will compare both theoretically and practically the different pursuits presented in this paper.

\subsection{Designing the Dictionaries} \label{Sec:ChoosingTheDictionaries}
We commence by describing the design of the dictionaries, and in the next subsection continue to the actual generation of the signals. In our experiments, the signal is one dimensional and therefore $m_0 = 1$. Moreover, for simplicity, the dictionary in every layer contains a single atom with its shifts and thus $m_i=1 \ \forall 1 \leq i \leq K$. We should note that the choice of a single atom simplifies the involved pursuit problem, but as we will see, even in such a case the suggested layered pursuits (including the forward pass) may fail. This is because the mutual coherence and the amount of non-zeros are still non-trivial.

In the first layer we choose this filter to be the analytically defined discrete Meyer Wavelet of length $n_0 = 29$. In order to obtain sparser representations and improve the coherence of the global dictionary $\D_1$, we employ a stride of $s_0 = 6$, resulting in $\mu(\D_1) = 2.44\times 10^{-4}$. As a consequence of our choice of $\D_1$, the signals resulting from our model are a superposition of shifted versions of discrete Meyer Wavelets, multiplied by different coefficients.

Recall that in the context of the layered thresholding algorithm, our theoretical study has shown that the stability of the pursuit depends on the ratio $|\Gamma_i^{\text{min}}|/|\Gamma_i^{\text{max}}|$. As such, in addition to requiring $\mu(\D_i)$ to be small, we would also like the ratio $|\Gamma_i^{\text{min}}|/|\Gamma_i^{\text{max}}|$ to be as close as possible to one. Since the sparse vectors satisfy $\Gama_{i} = \D_{i+1} \Gama_{i+1} \ \forall 1 \leq i \leq K-1$, one can control this ratio by forcing the entries in the dictionaries $\{\D_i\}_{i=2}^K$ to be\footnote{Note that we do not force the entries in $\D_1$ to be discrete since $\X=\D_1\Gama_1$ and the ratio of the entries in $\X$ is of no significance to the success of the layered thresholding algorithms.} discrete\footnote{In our experiments, the non-zero entries in the deepest representation $\Gama_K$ are chosen to be $\pm 1$. As such, the sparse vector $\Gama_{K-1} = \D_K \Gama_K$ is a superposition of filters (or their negative) taken from the dictionary $\D_K$. If the entries in $\D_K$ are non-discrete then the summation of two filters can result in extremely small values in $\Gama_{K-1}$, which in turn would lead to a very small $|\Gama_{K-1}^{\text{min}}|$ and a bad ratio. On the other hand, if the atoms are chosen to be discrete, this would not happen since the entries would simply cancel each other.}. Following this observation, and motivated by the benefits of a sparse dictionary, we generate a filter of length $20$ with $7$ non-zero entries belonging to the set $\left\{ -8,-7,...,7,8 \right\}$ (these are the entries before the atom is normalized to a unit $\ell_2$ norm). In practice, this is done by sampling random vectors satisfying these constraints and choosing one resulting in a good mutual coherence. For simplicity, all $\{\D_i\}_{i=2}^K$ are created from the very same local atom, i.e.  $n_i = 20 \ \forall 1 \leq i \leq K-1$. Moreover, in all the dictionaries this atom is shifted by a stride of $s_i = 6$, leading to $\mu(\D_i) = 4.33 \times 10^{-3}$. Note that in the above description the specific number of layers $K$ was purposely omitted, as this number will vary in the following experiments.

\subsection{Noiseless Experiments} \label{NoiselessExperiments}
We now move to the task of sampling a signal when the number of layers is $K=3$. First, we draw a random $\Gama_3$ of length $100$ with an $\ell_0$ norm in the range $\left[20,66\right]$ and set each non-zero coefficient in it to $\pm 1$, with equal probability. Given the dictionaries and the sampled sparse vector $\Gama_3$, we then compute the representations $\Gama_2$, $\Gama_1$ and the signal $\X$, which are of length $600$, $3,600$ and $N=21,600$, respectively. The obtained sparse vectors satisfy $\| \Gama_1 \|_{0,\infty}^\ss = 8$, $5 \leq \| \Gama_2 \|_{0,\infty}^\ss \leq 6$ and $ 3 \leq \| \Gama_3 \|_{0,\infty}^\ss \leq 7$.

Given the signals, we attempt to retrieve their underlying representations using the layered pursuits presented in this work. Recall that our analysis in Section \ref{Sec:TheoreticalStudy} and \ref{Sec:Future} indicates that the layered hard thresholding is superior to its soft counterpart, which is equivalent to the forward pass, and that the layered BP is even better than both of these algorithms. We now turn to asserting this claim empirically. While doing so, we aim to study the gap between the theoretical guarantees presented throughout our paper and the empirical performance obtained in practice.

\begin{figure}[!htbp]
	\centering
	\includegraphics[width=0.9\textwidth]{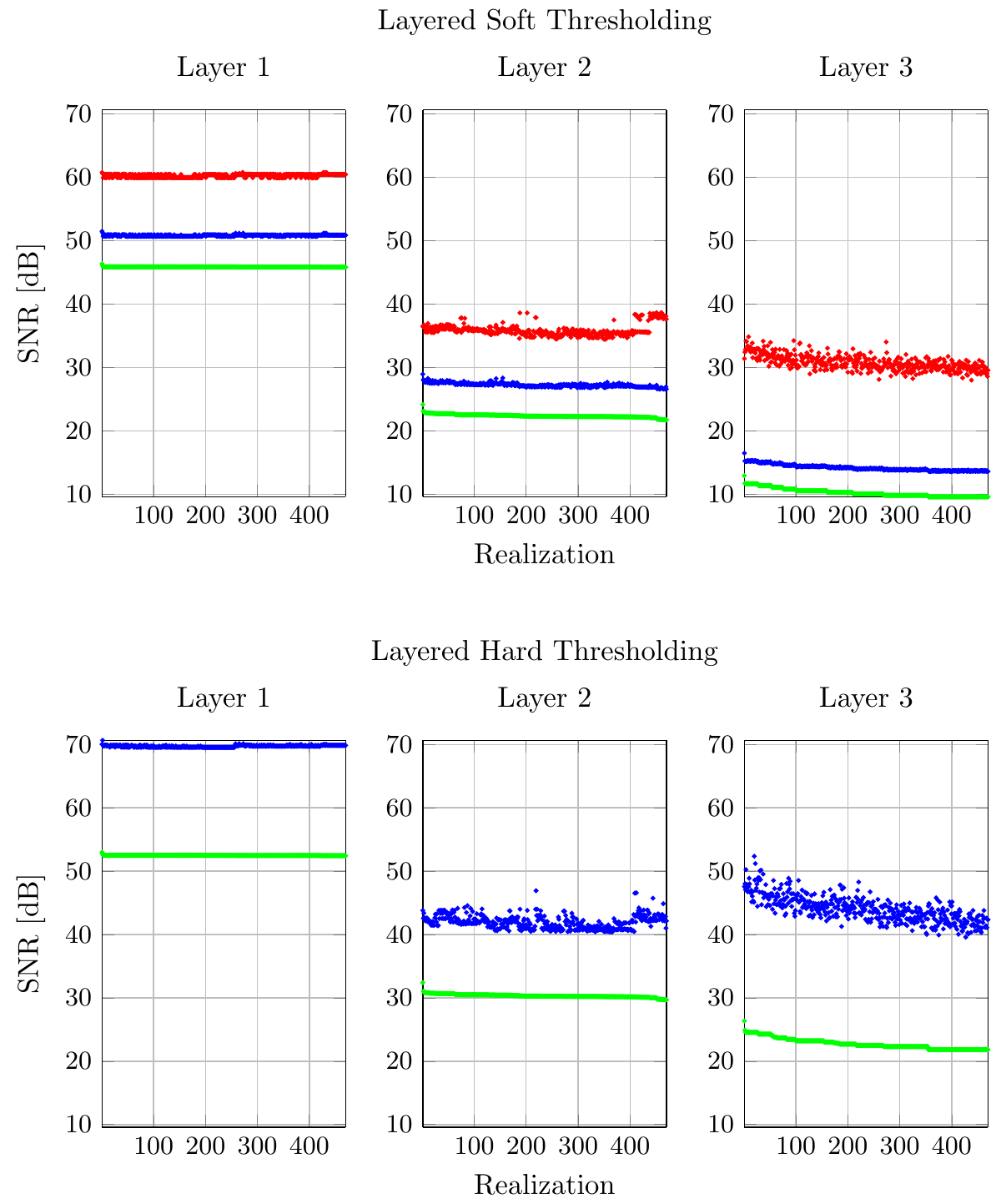}
	\caption{Comparison between the performance of the layered thresholding algorithms in a noiseless setting. All the signals presented here satisfy the assumptions of Theorem \ref{Thm:StabilityLayeredHardThresholding} and \ref{Thm:StabilityLayeredSoftThresholding} and indeed the correct support of all the representations are recovered. The horizontal axis plots the realization number, while the vertical axis shows the SNR (the higher the better). Blue: layered thresholding algorithm with the theoretically justifiable thresholds. Green: the theoretical bound on the error. Red: layered soft thresholding algorithm with oracle thresholds. Note that the points are sorted according to the theoretical bound.}
	\label{Fig:Noiseless}
\end{figure}

For every signal $\X$ (termed {\em realization} below), we employ the layered hard thresholding algorithm. The thresholds are set to be the ones presented in Theorem \ref{Thm:StabilityLayeredHardThresholding}, since the $\Loi$ norms of the representations of each $\X$ satisfy the assumptions of this theorem. Given the estimated sparse vectors $\{\hat{\Gama}_i\}_{i=1}^3$, we then compute the errors $\| \hat{\Gama}_i - \Gama_i \|_{2,\infty}^\pp$ and compare these to the theoretical bounds. While doing so, we also verify that the correct support is indeed retrieved, as our theorem guarantees. Next, the same process is repeated for the layered soft thresholding algorithm, with the exception that the thresholds and the bound on the distance are computed according to Theorem \ref{Thm:StabilityLayeredSoftThresholding}. We note that the assumptions of this hold as well for every signal $\X$. The results for both algorithms are depicted in Figure \ref{Fig:Noiseless} in terms of the local signal to noise ratio (SNR), defined as $20 \log 10 \Big(\frac{\| \Gama_i \|_{2,\infty}^\pp}{\| \hat{\Gama}_i - \Gama_i \|_{2,\infty}^\pp}\Big)$. Due to the locality of the analysis, we choose to deviate from the classical definition of the (global) SNR, given by $20 \log 10 \left(\frac{\| \Gama_i \|_2}{\| \hat{\Gama}_i - \Gama_i \|_2}\right)$.

Several remarks are due here. First and foremost, the theoretical bounds indeed hold, since the blue points are above their corresponding green ones and the correct supports are always recovered. Second, our analysis predicts that the distance between the estimated sparse representation, $\hat{\Gama}_i$, and the true ones, $\Gama_i$, should increase with the layer. This is evident by the decrease in the values of the green points with the layers. The empirical results presented here (blue dots) corroborate this prognosis, as the error in both algorithms is lowest in the first layer and highest in the last\footnote{Interestingly, the error in the layered hard thresholding algorithm is approximately equal in the second and third layers.}. Third, our analysis suggests that the layered hard thresholding algorithm should be superior to its soft counterpart. Once again, this can be deduced from the figure by comparing the values of the green points in both of the algorithms. The empirical results presented in \mbox{Figure \ref{Fig:Noiseless}} confirm this behavior, as can be clearly seen by comparing the errors (blue points) obtained by both algorithms in the $i$-th layer. One should note that the performance gap exhibited here is due to the constant $\beta_i$ being subtracted from every entry in the soft thresholding algorithm.

The implications of the above discussion might be troubling in the context of CNN, as what this experiment shows is a deterioration of the \textbf{empirical} SNR throughout the layers of the network. Is this truly the behavior of CNN? Recall that in practice the biases of the different layers (thresholds) are learned in order to achieve the best possible performance in solving a certain task. As such, it might be possible that the decline in SNR presented here is alleviated when better thresholds are employed in lieu of the theoretical ones used thus far. We demonstrate this by running the layered soft thresholding algorithm with an oracle parameter, chosen to be the minimal threshold that leads to $\| \Gama_i \|_0$ non-zeros being chosen in the estimated sparse representation $\hat{\Gama}_i$. The results for this are presented in \mbox{Figure \ref{Fig:Noiseless}} and colored in red. Indeed, we observe that this better choice of parameters improves the empirical performance of the layered soft thresholding algorithm and leads to a slower decline in SNR. Still, the performance of the layered soft thresholding is inferior to that of its hard variant\footnote{Note that in the layered hard thresholding, as long as the correct support is chosen, the threshold does not affect the error and as such the oracle version for it is meaningless.}, as can be seen by comparing the red points with the blue ones in the subplots below.

Next, we proceed our experiments by running the layered BP algorithm, as defined in Section \ref{Sec:Future}, on the same set of signals. Recall that one of the prime motivations for proposing this algorithm was its ability to retrieve the exact underlying representations, as justified theoretically in Theorem \ref{Thm:SuccessLayeredBP}. In our experiments, we validate this claim by checking that its conditions hold for each signal and that the underlying representations are indeed retrieved. We omit showing a plot for this and comparing it to the layered thresholding algorithms since the errors obtained are simply zeros.

\subsection{Noisy Experiments}
Having established the stability of our proposed algorithms in a perfect scenario, where $\epsilon_0 = 0$, we now turn to a noisy setting. Naturally, the estimation task becomes now even more challenging -- not only does the SNR drop with each layer, as demonstrated previously, but also the input SNR is no longer infinity. In order to facilitate the success of our algorithms, in this section we demonstrate the empirical performance and theoretical bounds on $K=2$ layers and a small noise level.

\begin{figure}[!htbp]
	\centering
	\includegraphics[width=0.9\textwidth]{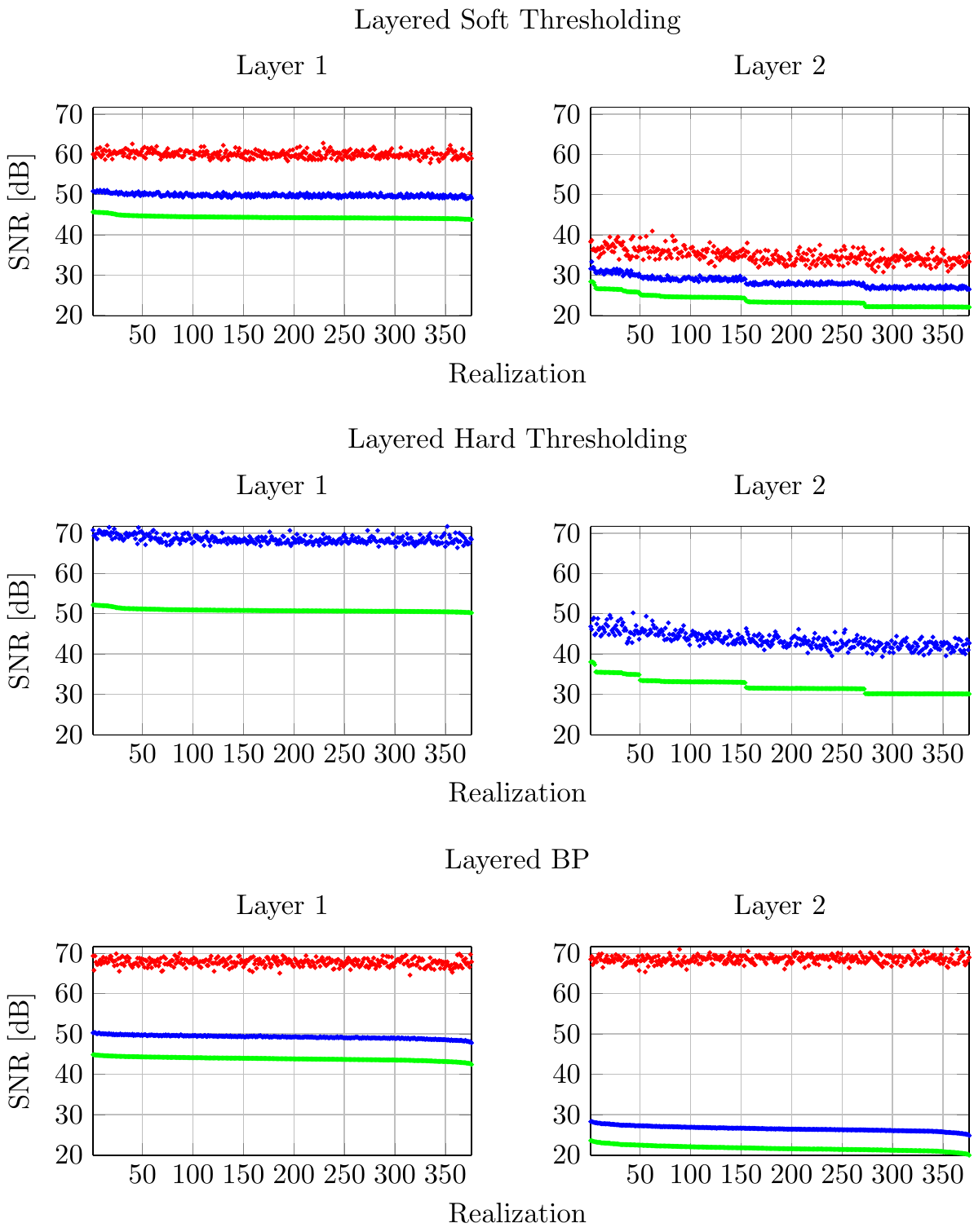}
	\caption{Comparison between the performance of the layered pursuit algorithms in a noisy setting. All the signals presented here satisfy the assumptions of Theorem \ref{Thm:StabilityLayeredHardThresholding}, \ref{Thm:StabilityLayeredSoftThresholding} and \ref{Thm:StabilityLayeredBP} and the correct support of all the representations are indeed recovered. The horizontal axis plots the realization number, while the vertical axis shows the SNR (the higher the better). Blue: layered pursuits with the theoretically justifiable thresholds. Green: the theoretical bound on the error. Red: layered soft thresholding with oracle thresholds and layered BP with hand-picked parameters. Note that the points are sorted according to the theoretical bound.}
   \label{Fig:Noisy}
\end{figure}

Similar to the previous subsection, we begin by sampling a signal $\X$. To this end, we draw a random $\Gama_2$ of length $100$ where $20 \leq \| \Gama_2 \|_0 \leq 66$. Each non-zero coefficient in it is then set to $\pm 1$, with equal probability. Given the dictionaries and the sampled sparse vector $\Gama_2$, we then compute the representation $\Gama_1$ and the signal $\X$, which are of length $600$ and $3,600$, respectively. The $\Loi$ norm of the obtained sparse vectors satisfies $7 \leq \| \Gama_1 \|_{0,\infty}^\ss \leq 8$ and $3 \leq \| \Gama_2 \|_{0,\infty}^\ss \leq 7$.

Next, we contaminate each signal $\X$ with a zero-mean white additive Gaussian noise $\E$, creating a signal $\Y = \X + \E$. The average SNR of the obtained noisy signals is \mbox{$68.53$ dB.} These are then fed into the layered pursuits, resulting in a set of estimated sparse representations, $\hat{\Gama}_i$. We note that the $\Loi$ norms of the representations of each $\X$ satisfy the assumptions of Theorem \ref{Thm:StabilityLayeredHardThresholding}, \ref{Thm:StabilityLayeredSoftThresholding} and \ref{Thm:StabilityLayeredBP}. As such, the parameters for every algorithm are chosen according to our theoretical study. For each estimated representation we compute the error $\| \hat{\Gama}_i - \Gama_i \|_{2,\infty}^\pp$ and its corresponding theoretical bound obtained from the aforementioned theorems. Since the underlying representations satisfy the assumptions of the stability theorem for the layered thresholding algorithms, for each signal we verify that indeed the correct support is found. As for the layered BP, our stability analysis guarantees that the support retrieved should be contained in the true one and coefficients that are large enough in $\Gama_i$ should be retrieved. In practice, the layered BP always finds the full support.

We present the obtained results in terms of the local SNR in Figure \ref{Fig:Noisy}, showing the stability of the different algorithms that is in accordance with our theoretical bounds. Similar to the noiseless experiment, we observe that for all the algorithms the error increases both theoretically (green points) and empirically (blue points) with the layer depth. As previously discussed, a performance gap exists between the soft and hard layered thresholding algorithms. To mitigate this, we run the layered soft thresholding with an oracle parameter and compare the obtained errors (red points) to those of the other algorithms. The results, depicted in the same figure, show a clear improvement in the performance.

Interestingly, although theoretically superior, the layered BP leads to similar performance to that of the layered soft thresholding and worse performance than that of the layered hard thresholding (when comparing the blue points). We attribute this phenomenon to the suboptimal choice of the parameter $\xi_i$, which was chosen thus far according to our theoretical analysis. To validate this suspicion, we run the layered BP with hand-picked $\xi_i$ and plot the obtained SNR in red in Figure \ref{Fig:Noisy}. Not only are the correct supports retrieved for all the signals, but we can also see a clear improvement in terms of the SNR. In the first layer, the layered BP outperforms the layered soft thresholding and leads to similar results to those of the layered hard thresholding, while in the second, the layered BP significantly outperforms both of the other pursuit algorithms.

Thus far, our experiments focused on a setting where the ratio of the coefficients in $\Gama_i$ is reasonable. One should note, however, that the superiority of the layered BP becomes conspicuous once this ratio is spoiled. In this case, the layered thresholding algorithms will fail, while the layered BP will still succeed. To illustrate this, we create a signal using the dictionaries delineated in subsection \ref{Sec:ChoosingTheDictionaries}, where the number of layers is $K=5$. We first draw a random $\Gama_5$ of length $100$ where its $\ell_0$ norm is in the range $20 \leq \| \Gama_5 \|_0 \leq 66$, and then set the non-zero coefficients in $\Gama_5$, similar to how it was done in the previous experiments. Given the dictionaries and the sampled sparse vector $\Gama_5$, we compute the representation $\{\Gama_i\}_{i=1}^4$ and the signal $\X$, which is of length $N=777,600$. The $\Loi$ norms of the obtained sparse vectors are $\| \Gama_1 \|_{0,\infty}^\ss = 8$, $\| \Gama_2 \|_{0,\infty}^\ss = 6$, $\| \Gama_3 \|_{0,\infty}^\ss = 6$, $\| \Gama_4 \|_{0,\infty}^\ss = 6$ and $4 \leq \| \Gama_5 \|_{0,\infty}^\ss \leq 7$. Besides the depth of the network, the main difference between this experiment and the previous ones is the coefficient ratio. While the ratio of the deepest representation $\Gama_5$ is equal to $1$, due to the coefficients in it being equal to $\pm 1$, the ratio of $\Gama_1$ is equal to $2.44 \times 10^{-4}$. As a consequence, the theoretical results we have presented for the layered thresholding algorithms do not hold, while those of the layered BP still do.

Next, each signal $\X$ is contaminated with a zero-mean white additive Gaussian noise $\E$, resulting in a noisy signal $\Y = \X + \E$. The average SNR of the noisy signals obtained is $124.43$ dB. Note that this is a weak noise, chosen due to the deterioration of the SNR throughout the layers (one that is worsened when the theoretical parameters are employed). The signals are then fed into the layered BP algorithm, resulting in a set of estimated sparse representations, $\hat{\Gama}_i$. The parameters $\xi_i$ employed are the theoretically justified ones, $\xi_i = 4 \epsilon_{i-1}$. We should note that in our experiments we attempted to run the layered thresholding algorithms, however, as our theory predicts these failed in recovering the correct supports.

\begin{figure}[t]
	\centering
	\includegraphics[width=0.9\textwidth]{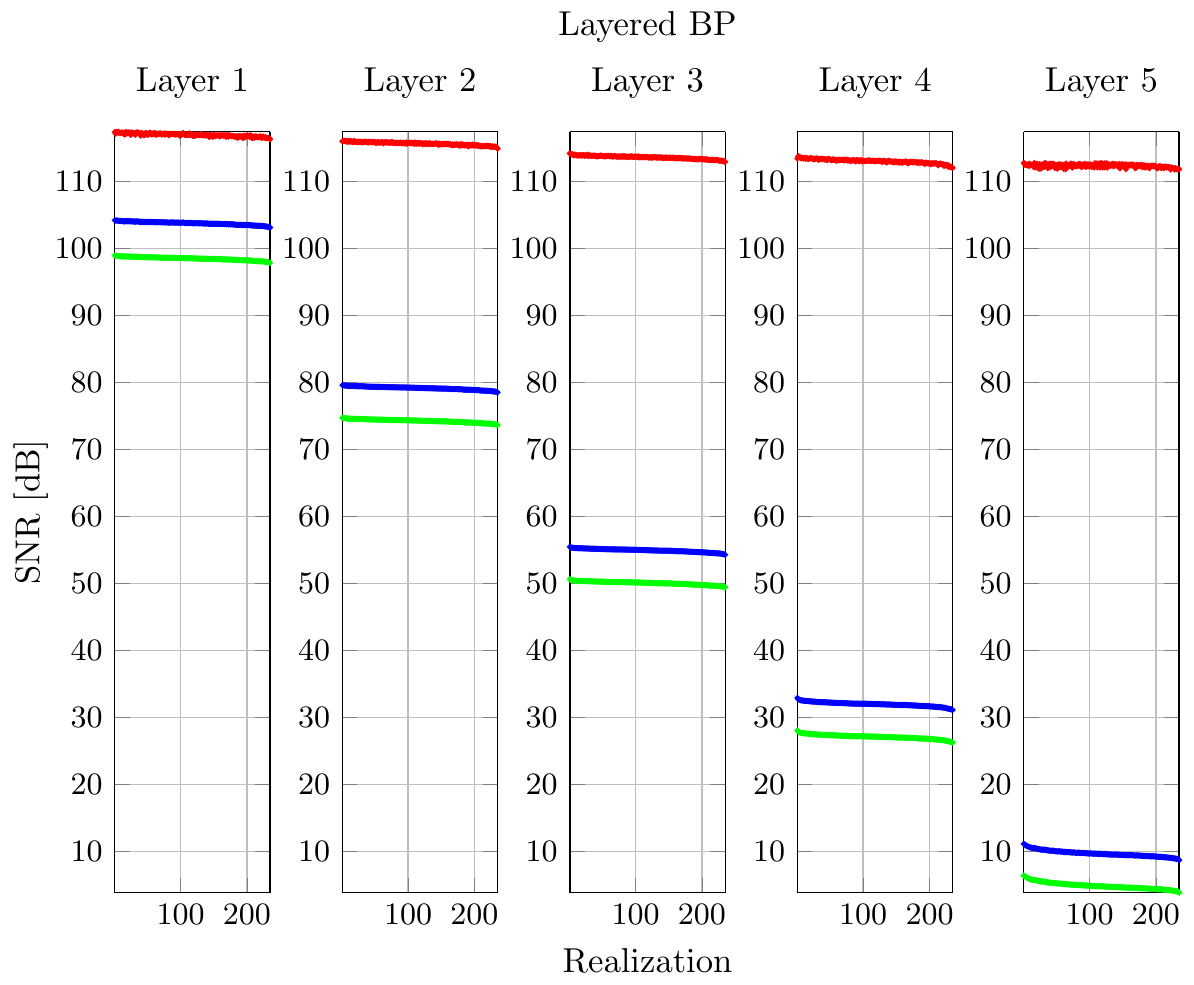}
	\caption{The performance of the layered BP algorithm in a noisy setting. All the signals presented here satisfy the assumptions of Theorem \ref{Thm:StabilityLayeredBP} and the correct support of all the representations are recovered. The horizontal axis plots the realization number, while the vertical axis shows the SNR (the higher the better). Blue: layered BP with the theoretically justifiable thresholds. Green: the theoretical bound on the error. Red: layered BP with hand-picked parameters. Note that the points are sorted according to the theoretical bound.}
   \label{Fig:BP}
\end{figure}

Given the estimated representations, we compute the errors $\| \hat{\Gama}_i - \Gama_i \|_{2,\infty}^\pp$ and compare these to their corresponding theoretical bounds, obtained from Theorem \ref{Thm:StabilityLayeredBP}. In addition, we verify that the retrieved supports are contained in the true one, as the theorem guarantees. In practice, we obtain that the layered BP always finds the full support. The obtained results are depicted in Figure \ref{Fig:BP} in terms of the local SNR. For comparison, we run the layered BP with hand-picked $\xi_i$ and present the obtained results in the same figure. We conclude that the layered BP remains stable despite the poor coefficient ratio, unlike the layered thresholding algorithms. Moreover, tuning the $\xi_i$ results in a much better performance, similar to what we have seen in the previous experiment.

At this point, one might ponder as to whether the hurdle of poor coefficient ratio is one that the layered soft thresholding (forward pass) can not overcome. We believe that several ideas currently used in CNN, such as Batch Normalization \citep{ioffe2015batch} or Local Response Normalization \citep{krizhevsky2012imagenet}, are tightly connected to this problem. However, their exact relation to this issue and its theoretical analysis is a matter of future work.

\section{Conclusion} \label{Sec:Conclusion}
Definition: ``A guiding question is the fundamental query that directs the search for understanding'' \citep{traver1998good}. In this work our guiding question was who are the signals that the CNN architecture is designed for? To answer this we have defined the ML-CSC model, for which the thresholding pursuit is nothing but the forward pass of the CNN. Although nothing promises that the forward pass will lead to the original representation of a signal emerging from the ML-CSC model, we have shown this is indeed the case. Having established the relevance of our model to CNN, we then turned to its theoretical analysis. In particular, we provided guarantees for the uniqueness of the feature maps CNN aims to recover, and the stability of the problem CNN aims to solve.

Inspired by the evolution of the pursuit methods in the theory of Sparse-Land, we continued our work by proposing the layered BP algorithm. In the noiseless case, this was theoretically shown to be capable of finding the unique solution of the deep coding problem, the existence of which has been also guaranteed; while in the noisy setting, we have proved the stability of this algorithm.

We analyzed the theoretical benefits of two popular ideas employed in the CNN community, namely the use of sparse filters and the spatial-stride. Leveraging those, we then generated signals satisfying the ML-CSC assumptions and demonstrated the performance of the pursuits presented throughout this work.

We conclude this work by presenting our ongoing research directions:
\begin{enumerate}
\item Through this paper we have assumed the worst -- an adversary noise. Can our theoretical analysis be extended to a setting where the noise is random?
\item Thus far in tackling the deep coding problem, we have restricted ourself to existing methods, such as the forward pass of the CNN or deconvolutional networks \citep{zeiler2010deconvolutional}. Can we suggest better approximations for the solution of this problem?
\item Clearly a relation exists between our proposed layered iterative thresholding algorithm and the current throne holder in the task of image recognition -- residual networks \citep{he2015deep}. Can our theory reveal the benefits of introducing skip connections to a CNN?
\item What is the role of common tricks currently employed in CNN in the context of the ML-CSC model? These include but are not limited to, Batch Normalization \citep{ioffe2015batch}, Local Response Normalization \citep{krizhevsky2012imagenet}, Dropout \citep{srivastava2014dropout} and Pooling \citep{le1990handwritten,krizhevsky2012imagenet,simonyan2014very}.
\end{enumerate}


\acks{The research leading to these results has received funding from the European Research Council under European Union's Seventh Framework Programme, ERC Grant agreement no. 320649. The authors would like to thank Jeremias Sulam for the inspiring discussions and creative advice.}

\appendix
\renewcommand*{\theHsection}{chX.\the\value{section}}

\section{\\Uniqueness via the Mutual Coherence (Proof of Theorem \ref{Thm:Uniqueness})} \label{App:uniqueness}
\begin{proof}
In \citep{Papyan2016_1} a solution $\Gama$ to the $\Poi$ problem, as defined in Equation \eqref{Eq:Poi}, was shown to be unique assuming that $\| \Gama \|_{0,\infty}^\ss < \frac{1}{2} \left( 1 + \frac{1}{\mu(\D)} \right)$. In other words, if the true representation is sparse enough in the $\Loi$ sense, no other solution is possible. Herein, we leverage this claim in order to prove the uniqueness of the $\DCP$ problem.

Let $\{\hat{\Gama}_i\}_{i=1}^{K}$ be a set of representations of the signal $\X$, obtained by solving the $\DCP$ problem. According to our assumptions, $\| \Gama_1 \|_{0,\infty}^\ss < \frac{1}{2} \left( 1 + \frac{1}{\mu(\D_1)} \right)$. Moreover, since the set $\{\hat{\Gama}_i\}_{i=1}^{K}$ is a solution of the $\DCP$ problem, we also have that $\| \hat{\Gama}_1 \|_{0,\infty}^\ss \leq \lambda_1 < \frac{1}{2} \left( 1 + \frac{1}{\mu(\D_1)} \right)$. As such, in light of the aforementioned uniqueness theorem, both representations are equal. Once we have concluded that $\Gama_1 = \hat{\Gama}_1$, we would also like to show that the representations $\Gama_2$ and $\hat{\Gama}_2$ are identical. Similarly, the assumptions $\| \Gama_2 \|_{0,\infty}^\ss < \frac{1}{2} \left( 1 + \frac{1}{\mu(\D_i)} \right)$ and $\| \hat{\Gama}_2 \|_{0,\infty}^\ss \leq \lambda_2 < \frac{1}{2} \left( 1 + \frac{1}{\mu(\D_2)} \right)$ guarantee that $\Gama_2 = \hat{\Gama}_2$. The same set of steps can be applied for all $1\leq i\leq K$, leading to the fact that both sets of representations are identical.
\end{proof}

\section[\texorpdfstring{Global Stability of the $\DCPE$ Problem (Proof of Theorem \ref{Thm:GlobalStability})}{Global Stability of the DCP Problem (Proof of Theorem \ref{Thm:GlobalStability})}]{\\Global Stability of the $\DCPE$ Problem (Proof of Theorem \ref{Thm:GlobalStability})} \label{App:GlobalStability}

\begin{proof}
In \citep{Papyan2016_2}, for a signal $\Y = \X + \E = \D_1\Gama_1 + \E$, it was shown that if the following hold:
\begin{enumerate}
\itemsep0em
\item $\| \Gama_1 \|_{0,\infty}^\ss < \frac{1}{2} \left( 1 + \frac{1}{\mu(\D_1)} \right)$ and $\| \E \|_2 = \| \Y - \D_1\Gama_1 \|_2 \leq \mathcal{E}_0$,
\item $\| \hat{\Gama}_1 \|_{0,\infty}^\ss < \frac{1}{2} \left( 1 + \frac{1}{\mu(\D_1)} \right)$ and $\| \Y - \D_1\hat{\Gama}_1 \|_2 \leq \mathcal{E}_0$, \label{Eq:StabilityItem}
\end{enumerate}
then
\begin{equation}
\|\Delt_1\|_2^2=\|\Gama_1-\hat{\Gama}_1\|_2^2\leq\frac{4{\mathcal{E}_0}^2}{1-(2\|\Gama_1\|_{0,\infty}^\ss-1)\mu(\D_1)}={\mathcal{E}_1}^2.
\end{equation}
In the above, we have defined $\Delt_1$ as the difference between the true sparse vector, $\Gama_1$, and the corresponding representation obtained by solving the $\DCPE$ problem, $\hat{\Gama}_1$. In item \ref{Eq:StabilityItem} we have used the fact that the solution for the $\DCPE$ problem, $\hat{\Gama}_1$, must satisfy $\| \Y - \D_1\hat{\Gama}_1 \|_2 \leq \mathcal{E}_0$ and $\| \hat{\Gama}_1 \|_{0,\infty}^\ss \leq \lambda_1$; and our assumption that $\lambda_1<\frac{1}{2} \left( 1 + \frac{1}{\mu(\D_1)} \right)$.
Next, notice that $\hat{\Gama}_1 = \Gama_1 + \Delt_1 = \D_2\Gama_2 + \Delt_1$, and that the following hold:
\begin{enumerate}
\itemsep0em
\item $\| \Gama_2 \|_{0,\infty}^\ss < \frac{1}{2} \left( 1 + \frac{1}{\mu(\D_2)} \right)$ and $\| \Delt_1 \|_2 = \| \hat{\Gama}_1 - \D_2\Gama_2 \|_2 \leq \mathcal{E}_1$,
\item $\| \hat{\Gama}_2 \|_{0,\infty}^\ss < \frac{1}{2} \left( 1 + \frac{1}{\mu(\D_2)} \right)$ and $\| \hat{\Gama}_1 - \D_2\hat{\Gama}_2 \|_2 \leq \mathcal{E}_1$.
\end{enumerate}
The second item relies on the fact that both $\hat{\Gama}_1$ and $\hat{\Gama}_2$, obtained by solving the $\DCPE$ problem, must satisfy $\| \hat{\Gama}_1 - \D_2\hat{\Gama}_2 \|_2 \leq \mathcal{E}_1$ and $\| \hat{\Gama}_2 \|_{0,\infty}^\ss \leq \lambda_2$. In addition, the second expression uses the assumption that $\lambda_2 < \frac{1}{2} \left( 1 + \frac{1}{\mu(\D_2)} \right)$. Employing once again the aforementioned stability theorem, we are guaranteed that
\begin{equation}
\|\Delt_2\|_2^2=\|\Gama_2-\hat{\Gama}_2\|_2^2\leq\frac{4{\mathcal{E}_1}^2}{1-(2\|\Gama_2\|_{0,\infty}^\ss-1)\mu(\D_2)}={\mathcal{E}_2}^2.
\end{equation}
Using the same set of steps presented above, we conclude that
\begin{equation}
\forall \ 1\leq i\leq K \qquad \| \Gama_i- \hat{\Gama}_i \|_2^2 \leq {\mathcal{E}_i}^2,
\end{equation}
as claimed.
\end{proof}

\section{\\Stable Recovery of Hard Thresholding in the Presence of Noise (Proof of Lemma \ref{Thm:StabilityHardThresholding})}
\label{App:StabilityHardThresholding}
\begin{proof}
Denote by $\mathcal{T}_1$ the support of $\Gama_1$. Denote further the $i$-th atom from $\D_1$ by $\d_{1,i}$. The success of the hard thresholding algorithm with threshold $\beta_1$ in recovering the correct support is guaranteed if the following holds
\begin{equation}
\min_{i\in\mathcal{T}_1} \left| \d_{1,i}^T \Y \right| > \beta_1 > \max_{j\notin\mathcal{T}_1} \left| \d_{1,j}^T \Y \right|.
\end{equation}
Using the same set of steps as those used in proving Theorem 4 in \citep{Papyan2016_2}, we can lower bound the left-hand-side by
\begin{equation}
\min_{i\in\mathcal{T}_1} \left| \d_{1,i}^T \Y \right|
\geq |\Gama_1^{\text{min}}| - ( \| \Gama_1 \|_{0,\infty}^\ss - 1 ) \mu(\D_1) |\Gama_1^{\text{max}}| - \epsilon_0
\end{equation}
and upper bound the right-hand-side via
\begin{equation}
\| \Gama_1 \|_{0,\infty}^\ss \mu(\D_1) |\Gama_1^{\text{max}}| + \epsilon_0
\geq \max_{j\notin\mathcal{T}_1} \left| \d_{1,j}^T \Y \right|.
\end{equation}
Next, by requiring
\begin{align} \label{Eq:ConditionStabilityThresholding}
\min_{i\in\mathcal{T}_1} \left| \d_{1,i}^T \Y \right|
\geq & |\Gama_1^{\text{min}}| - ( \| \Gama_1 \|_{0,\infty}^\ss - 1 ) \mu(\D_1) |\Gama_1^{\text{max}}| - \epsilon_0 \\
> & \ \beta_1 \\
> & \| \Gama_1 \|_{0,\infty}^\ss \mu(\D_1) |\Gama_1^{\text{max}}| + \epsilon_0 \\
\geq & \max_{j\notin\mathcal{T}_1} \left| \d_{1,j}^T \Y \right|,
\end{align}
we ensure the success of the thresholding algorithm. This condition can be equally written as
\begin{equation}
\| \Gama_1 \|_{0,\infty}^\ss < \frac{1}{2} \left( 1 + \frac{1}{\mu(\D_1)} \frac{ |\Gamma_1^{\text{min}}| }{ |\Gamma_1^{\text{max}}| } \right) - \frac{1}{\mu(\D_1)}\frac{\epsilon_0}{|\Gamma_1^{\text{max}}|}.
\end{equation}
Equation \eqref{Eq:ConditionStabilityThresholding} also implies that the threshold $\beta_1$ that should be employed must satisfy
\begin{equation} \label{Eq:ConditionBeta1}
|\Gama_1^{\text{min}}| - ( \| \Gama_1 \|_{0,\infty}^\ss - 1 ) \mu(\D_1) |\Gama_1^{\text{max}}| - \epsilon_0
> \beta_1
> \| \Gama_1 \|_{0,\infty}^\ss \mu(\D_1) |\Gama_1^{\text{max}}| + \epsilon_0.
\end{equation}

Thus far, we have considered the successful recovery of the support of $\Gama_1$. Next, assuming this correct support was recovered, we shall dwell on the deviation of the thresholding result, $\hat{\Gama}_1$, from the true $\Gama_1$. Denote by $\Gama_{1,\mathcal{T}_1}$ and $\hat{\Gama}_{1,\mathcal{T}_1}$ the vectors $\Gama_1$ and $\hat{\Gama}_1$ restricted to the support $\mathcal{T}_1$, respectively. We have that
\begin{align}
\| \Gama_1 - \hat{\Gama}_1 \|_\infty
& = \| \Gama_{1,\mathcal{T}_1} - \hat{\Gama}_{1,\mathcal{T}_1} \|_\infty \\
& = \left\| \left( \D_{1,\mathcal{T}_1}^T\D_{1,\mathcal{T}_1} \right)^{-1}\D_{1,\mathcal{T}_1}^T \X - \D_{1,\mathcal{T}_1}^T \Y \right\|_\infty \\
& = \left\| \left( \left( \D_{1,\mathcal{T}_1}^T\D_{1,\mathcal{T}_1} \right)^{-1}\D_{1,\mathcal{T}_1}^T - \D_{1,\mathcal{T}_1}^T \right) \X - \D_{1,\mathcal{T}_1}^T (\Y - \X) \right\|_\infty,
\end{align}
where the Gram $\D_{1,\mathcal{T}_1}^T\D_{1,\mathcal{T}_1}$ is invertible according to Lemma 1 in \citep{Papyan2016_1}. Using the triangle inequality of the $\ell_\infty$ norm and the relation $\X = \D_{1,\mathcal{T}_1}\Gama_{1,\mathcal{T}_1}$, we obtain
\begin{align*}
\| \Gama_1 - \hat{\Gama}_1 \|_\infty
& \leq \left\| \left( \left( \D_{1,\mathcal{T}_1}^T\D_{1,\mathcal{T}_1} \right)^{-1}\D_{1,\mathcal{T}_1}^T - \D_{1,\mathcal{T}_1}^T \right) \D_{1,\mathcal{T}_1}\Gama_{1,\mathcal{T}_1} \right\|_\infty + \Big\| \D_{1,\mathcal{T}_1}^T (\Y - \X) \Big\|_\infty \\
& = \left\| \left( \mathbf{I} - \D_{1,\mathcal{T}_1}^T\D_{1,\mathcal{T}_1} \right) \Gama_{1,\mathcal{T}_1} \right\|_\infty + \left\| \D_{1,\mathcal{T}_1}^T (\Y - \X) \right\|_\infty,
\end{align*}
where $\mathbf{I}$ is an identity matrix. Relying on the definition of the induced $\ell_\infty$ norm, the above is equal to
\begin{equation} \label{Eq:Delta1}
\| \Gama_1 - \hat{\Gama}_1 \|_\infty
\leq \left\| \mathbf{I} - \D_{1,\mathcal{T}_1}^T\D_{1,\mathcal{T}_1} \right\|_\infty \cdot \| \Gama_{1,\mathcal{T}_1} \|_\infty + \left\| \D_{1,\mathcal{T}_1}^T (\Y - \X) \right\|_\infty.
\end{equation}
In what follows, we shall upper bound both of the expressions in the right hand side of the inequality.

Beginning with the first term in the above inequality, $\left\| \mathbf{I} - \D_{1,\mathcal{T}_1}^T\D_{1,\mathcal{T}_1} \right\|_\infty$, recall that the induced infinity norm of a matrix is equal to its maximum absolute row sum. The diagonal entries of $\mathbf{I} - \D_{1,\mathcal{T}_1}^T\D_{1,\mathcal{T}_1}$ are equal to zero, due to the normalization of the atoms, while the off diagonal entries can be bounded by relying on the locality of the atoms and the definition of the $\Loi$ norm. As such, each row has at most $\| \Gama_1 \|_{0,\infty}^\ss-1$ non-zeros, where each is bounded by $\mu(\D_1)$ based on the definition of the mutual coherence. We conclude that the maximum absolute row sum can be bounded by
\begin{equation} \label{Eq:IdentityMinusGram}
\| \D_{1,\mathcal{T}_1}^T\D_{1,\mathcal{T}_1} - \mathbf{I} \|_\infty \leq (\| \Gama_1 \|_{0,\infty}^\ss - 1)\mu(\D_1).
\end{equation}
Next, moving to the second expression, define $\RR_{1,i}\in\mathbb{R}^{n_0\times N}$ to be the operator that extracts a filter of length $n_0$ from $\d_{1,i}$. Consequently, the operator $\RR_{1,i}^T$ pads a local filter of length $n_0$ with zeros, resulting in a global atom of length $N$. Notice that, due to the locality of the atoms $\RR_{1,i}^T\RR_{1,i}\d_{1,i}=\d_{1,i}$. Using this together with the Cauchy-Schwarz inequality, the normalization of the atoms, and the local bound on the error $\| \Y - \X \|_{2,\infty}^\pp\leq\epsilon_0$, we have that
\begin{align} \label{Eq:LocalNoise}
\left\| \D_{1,\mathcal{T}_1}^T (\Y - \X) \right\|_\infty
& = \underset{i\in\mathcal{T}_1}{\max} \left| \d_{1,i}^T (\Y - \X) \right| \\
& = \underset{i\in\mathcal{T}_1}{\max} \left| \left(\RR_{1,i}\d_{1,i}\right)^T \RR_{1,i}(\Y - \X) \right| \\
& \leq \underset{i\in\mathcal{T}_1}{\max} \ \| \RR_{1,i}\d_{1,i} \|_2 \cdot \| \RR_{1,i}(\Y - \X) \|_2 \\
& \leq \ 1 \cdot \| \Y - \X \|_{2,\infty}^\pp \\
& \leq \epsilon_0.
\end{align}
In the second to last inequality we have used Definition \ref{Def:LtinfLoip}, denoting the maximal $\ell_2$ norm of a \emph{patch} extracted from $\Y - \X$ by $\| \Y - \X \|_{2,\infty}^\pp$. Plugging \eqref{Eq:IdentityMinusGram} and \eqref{Eq:LocalNoise} into Equation \eqref{Eq:Delta1}, and using the fact that $\| \Gama_{1,\mathcal{T}_1} \|_\infty = |\Gama_1^{\text{max}}|$, we obtain that
\begin{equation} \label{Eq:Delta_inf}
\| \Gama_1 - \hat{\Gama}_1 \|_\infty
\leq (\| \Gama_1 \|_{0,\infty}^\ss - 1) \mu(\D_1) |\Gama_1^{\text{max}}| + \epsilon_0.
\end{equation}
In the remainder of this proof we will localize the above bound into one that is posed in terms of patch-errors. Note that $\| \Gama_1 - \hat{\Gama}_1 \|_{2,\infty}^\pp$ is equal to the maximal energy of an $n_1 m_1$-dimensional patch taken from it, where the $i$-th patch can be extracted using the operator $\PP_{1,i}$. Relying on this and the relation $\| \V \|_2 \leq \sqrt{\| \V \|_0} \ \| \V \|_\infty$, we have that
\begin{align}
\| \Gama_1 - \hat{\Gama}_1 \|_{2,\infty}^\pp
& = \max_i \left\| \PP_{1,i} \left( \Gama_1 - \hat{\Gama}_1 \right) \right\|_2 \\
& \leq \max_i \sqrt{ \left\| \PP_{1,i} \left( \Gama_1 - \hat{\Gama}_1 \right) \right\|_0 } \ \left \| \PP_{1,i} \left( \Gama_1 - \hat{\Gama}_1 \right) \right\|_\infty.
\end{align}
Recalling that, based on Definition \ref{Def:LtinfLoip}, $\| \Gama_1 - \hat{\Gama}_1\|_{0,\infty}^\pp$ denotes the maximal number of non-zeros in a patch of length $n_1 m_1$ extracted from this vector, we obtain that
\begin{align}
\| \Gama_1 - \hat{\Gama}_1 \|_{2,\infty}^\pp
& \leq \sqrt{ \| \Gama_1 - \hat{\Gama}_1 \|_{0,\infty}^\pp } \ \| \Gama_1 - \hat{\Gama}_1 \|_\infty \\
& \leq \sqrt{ \| \Gama_1 \|_{0,\infty}^\pp } \ \| \Gama_1 - \hat{\Gama}_1 \|_\infty.
\end{align}
In the last inequality we have used the success of the first stage in recovering the correct support, resulting in $\|\Gama_1 - \hat{\Gama}_1\|_{0,\infty}^\pp \leq \|\Gama_1\|_{0,\infty}^\pp$. Plugging inequality \eqref{Eq:Delta_inf} into the above equation, we conclude that
\begin{align}
\| \Gama_1 - \hat{\Gama}_1\|_{2,\infty}^\pp \leq \sqrt{\| \Gama_1 \|_{0,\infty}^\pp}\Big(\epsilon_0 + \mu(\D_1) \left( \| \Gama_1 \|_{0,\infty}^\ss - 1 \right) |\Gamma_1^{\text{max}}| \Big),
\end{align}
as claimed.
\end{proof}

\section{\\Stability of the Layered Hard Thresholding in the Presence of Noise (Proof of Theorem \ref{Thm:StabilityLayeredHardThresholding})}
\label{App:StabilityLayeredHardThresholding}
\begin{proof}
The stability of the first stage of the layered hard thresholding algorithm is obtained from Lemma \ref{Thm:StabilityHardThresholding}. Denoting by $\Delt_1 = \hat{\Gama}_1 - \Gama_1$, notice that $\hat{\Gama}_1 = \Gama_1 + \Delt_1 = \D_2\Gama_2 + \Delt_1$. In other words, $\Gama_1$ is a signal that admits a convolutional sparse representation $\D_2\Gama_2$, which is contaminated with noise $\Delt_1$, resulting in $\hat{\Gama}_1$. Next, we would like to employ Lemma \ref{Thm:StabilityHardThresholding} for the signal $\hat{\Gama}_1 = \Gama_1 + \Delt_1 = \D_2\Gama_2 + \Delt_1$, with the local noise level
\begin{equation}
\| \Delt_1 \|_{2,\infty}^\pp \leq \sqrt{\| \Gama_1 \|_{0,\infty}^\pp}\Big(\epsilon_0 + \mu(\D_1) \left( \| \Gama_1 \|_{0,\infty}^\ss - 1 \right) |\Gamma_1^{\text{max}}| \Big) = \epsilon_1,
\end{equation}
to obtain the stability of the second stage. To this end, we require its conditions to hold; in particular, the $\Loi$ norm of $\Gama_2$ to obey
\begin{equation}
\| \Gama_2 \|_{0,\infty}^\ss < \frac{1}{2} \left( 1 + \frac{1}{\mu(\D_2)} \frac{ |\Gamma_2^{\text{min}}| }{ |\Gamma_2^{\text{max}}| } \right) - \frac{1}{\mu(\D_2)}\frac{ \epsilon_1 }{|\Gamma_2^{\text{max}}|},
\end{equation}
and the threshold $\beta_2$ to satisfy
\begin{equation}
|\Gama_2^{\text{min}}| - ( \| \Gama_2 \|_{0,\infty}^\ss - 1 ) \mu(\D_2) |\Gama_2^{\text{max}}| - \epsilon_1 
> \ \beta_2
> \| \Gama_2 \|_{0,\infty}^\ss \mu(\D_2) |\Gama_2^{\text{max}}| + \epsilon_1.
\end{equation}
Assuming the above hold, Lemma \ref{Thm:StabilityHardThresholding} guarantees that the support of $\hat{\Gama}_2$ is equal to that of $\Gama_2$, and also that
\begin{equation}
\| \Gama_2 - \hat{\Gama}_2 \|_{2,\infty}^\pp \leq \sqrt{\| \Gama_2 \|_{0,\infty}^\pp}\Big(\epsilon_1 + \mu(\D_2) \left( \| \Gama_2 \|_{0,\infty}^\ss - 1 \right) |\Gamma_2^{\text{max}}| \Big) = \epsilon_2.
\end{equation}
Using the same steps as above, we obtain the desired claim for all the remaining layers, assuming that
\begin{equation}
\| \Gama_i \|_{0,\infty}^\ss < \frac{1}{2} \left( 1 + \frac{1}{\mu(\D_i)} \frac{ |\Gamma_i^{\text{min}}| }{ |\Gamma_i^{\text{max}}| } \right) - \frac{1}{\mu(\D_i)}\frac{ \epsilon_{i-1} }{|\Gamma_i^{\text{max}}|}
\end{equation}
and that the thresholds $\beta_i$ are chosen to satisfy
\begin{equation}
|\Gama_i^{\text{min}}| - ( \| \Gama_i \|_{0,\infty}^\ss - 1 ) \mu(\D_i) |\Gama_i^{\text{max}}| - \epsilon_{i-1} 
> \ \beta_i
> \| \Gama_i \|_{0,\infty}^\ss \mu(\D_i) |\Gama_i^{\text{max}}| + \epsilon_{i-1}. \label{Eq:ConditionBeta_i}
\end{equation}
This completes our proof.
\end{proof}

\section{\\Stable Recovery of Soft Thresholding in the Presence of Noise (Proof of Lemma \ref{Thm:StabilitySoftThresholding})}
\label{App:StabilitySoftThresholding}
\begin{proof}
The success of the soft thresholding algorithm with threshold $\beta_1$ in recovering the correct support is guaranteed if the following holds
\begin{equation}
\min_{i\in\mathcal{T}_1} \left| \d_{1,i}^T \Y \right| > \beta_1 > \max_{j\notin\mathcal{T}_1} \left| \d_{1,j}^T \Y \right|.
\end{equation}
Since the soft thresholding operator chooses all atoms with correlations greater than $\beta_1$, the above implies that the true support $\mathcal{T}_1$ will be chosen. This condition is equal to that of the hard thresholding algorithm, and thus using the same steps as in Lemma \ref{Thm:StabilityHardThresholding}, we are guaranteed that the correct support will be chosen under Assumptions \eqref{Assumption:SoftThreshAssumption1} and \eqref{Assumption:SoftThreshAssumption2}.

The difference between the hard thresholding algorithm and its soft counterpart becomes apparent once we consider the estimated sparse vector. While the former estimates the non-zero entries in $\hat{\Gama}_1$ by computing $\D_{1,\mathcal{T}_1}^T\Y$, the latter subtracts or adds a constant $\beta_1$ from these, obtaining $\D_{1,\mathcal{T}_1}^T\Y - \beta_1\B$, where $\B$ is a vector of $\pm1$. As a result, the distance between the true sparse vector and the estimated one is given by
\begin{align}
\| \Gama_1 - \hat{\Gama}_1 \|_\infty
& = \left\| \Gama_{1,\mathcal{T}_1} - \hat{\Gama}_{1,\mathcal{T}_1} \right\|_\infty \\
& = \left\| \left( \D_{1,\mathcal{T}_1}^T\D_{1,\mathcal{T}_1} \right)^{-1}\D_{1,\mathcal{T}_1}^T \X - \left( \D_{1,\mathcal{T}_1}^T\Y - \beta_1\B \right) \right\|_\infty \\
& \leq \left\| \left( \D_{1,\mathcal{T}_1}^T\D_{1,\mathcal{T}_1} \right)^{-1}\D_{1,\mathcal{T}_1}^T \X - \D_{1,\mathcal{T}_1}^T\Y \right\|_\infty + \left\| \beta_1\B \right\|_\infty,
\end{align}
where in the last step we have used the triangle inequality for the $\ell_\infty$ norm. Notice that $\| \beta_1\B \|_\infty = \beta_1$, since $\beta_1$ must be positive according to Equation \eqref{Eq:ConditionBeta1}. Combining this together with the same steps as those used in proving Lemma \ref{Thm:StabilityHardThresholding}, the above can be bounded by
\begin{equation}
\| \Gama_1 - \hat{\Gama}_1 \|_{2,\infty}^\pp \leq \sqrt{\| \Gama_1 \|_{0,\infty}^\pp}\Big(\epsilon_0 + \mu(\D_1) \left( \| \Gama_1 \|_{0,\infty}^\ss - 1 \right) |\Gamma_1^{\text{max}}| + \beta_1 \Big),
\end{equation}
resulting in the desired claim.
\end{proof}

\section{\\Stability of the Layered BP Algorithm in the Presence of Noise (Proof of Theorem \ref{Thm:StabilityLayeredBP})}
\label{App:StabilityLayeredBP}
\begin{proof}
In \citep{Papyan2016_2}, for a signal $\Y = \X + \E = \D_1\Gama_1 + \E$, it was shown that if the following hold:
\begin{enumerate} [\quad a) ]
\item $\| \Y - \X \|_{2,\infty}^\pp \leq \epsilon_0$; and
\item $\| \Gama_1 \|_{0,\infty}^\ss < \frac{1}{3} \left( 1 + \frac{1}{\mu(\D_1)} \right)$,
\end{enumerate}
then the solution $\hat{\Gama}_1$ for the Lagrangian formulation of the BP problem with $\beta_1 = 4 \epsilon_0$ (see Equation \eqref{Eq:LagrangianBP}) satisfies that
\begin{enumerate}
\item The support of the solution $\hat{\Gama}_1$ is contained in that of $\Gama_1$;
\item $\| \Delt_1 \|_\infty = \| \Gama_1 - \hat{\Gama}_1 \|_\infty \leq 7.5 \ \epsilon_0$;
\item In particular, every entry of $\Gama_1$ greater in absolute value than $7.5 \ \epsilon_0$ is guaranteed to be recovered; and
\item The solution $\hat{\Gama}_1$ is the unique minimizer of the Lagrangian BP problem (Equation~\eqref{Eq:LagrangianBP}).
\end{enumerate}
Using similar steps to those employed in the proof of Theorem \ref{Thm:StabilityLayeredHardThresholding}, we obtain that
\begin{equation}
\| \Delt_1 \|_{2,\infty}^\pp \leq \sqrt{ \| \Delt_1 \|_{0,\infty}^\pp } \ \| \Delt_1 \|_\infty.
\end{equation}
Plugging above the inequality $\| \Delt_1 \|_\infty \leq 7.5 \ \epsilon_0$, we get
\begin{equation}
\| \Delt_1 \|_{2,\infty}^\pp \leq \sqrt{ \| \Delt_1 \|_{0,\infty}^\pp } \ 7.5 \ \epsilon_0.
\end{equation}
Since the support of $\hat{\Gama}_1$ is contained in that of $\Gama_1$, we have that $\| \Delt_1 \|_{0,\infty}^\pp = \| \Gama_1 - \hat{\Gama}_1 \|_{0,\infty}^\pp \leq \| \Gama_1 \|_{0,\infty}^\pp$, leading to
\begin{equation}
\| \Delt_1 \|_{2,\infty}^\pp \leq \sqrt{ \| \Gama_1 \|_{0,\infty}^\pp } \ 7.5 \ \epsilon_0 = \epsilon_1.
\end{equation}
We conclude that the first stage of the layered BP is stable and the following must hold
\begin{enumerate}
\item The support of the solution $\hat{\Gama}_1$ is contained in that of $\Gama_1$;
\item $\| \Delt_1 \|_{2,\infty}^\pp \leq \ \epsilon_1$;
\item In particular, every entry of $\Gama_1$ greater in absolute value than $\frac{\epsilon_1}{\sqrt{ \| \Gama_{1} \|_{0,\infty}^\pp }}$ is guaranteed to be recovered; and
\item The solution $\hat{\Gama}_1$ is the unique minimizer of the Lagrangian BP problem (Equation~\eqref{Eq:LagrangianBP}).
\end{enumerate}

Next, we turn to the stability of the second stage of the layered BP algorithm. Notice that $\hat{\Gama}_1 = \Gama_1 + \Delt_1 = \D_2\Gama_2 + \Delt_1$. Put differently, $\Gama_1$ is a signal that admits a convolutional sparse representation $\D_2\Gama_2$ that is perturbed by $\Delt_1$, resulting in $\hat{\Gama}_1$. As such, we can invoke once again the same theorem from \citep{Papyan2016_2}. Since we have that
\begin{enumerate} [\quad a) ]
\item $\| \Delt_1 \|_{2,\infty}^\pp \leq \epsilon_1$; and
\item $\| \Gama_2 \|_{0,\infty}^\ss < \frac{1}{3} \left( 1 + \frac{1}{\mu(\D_2)} \right)$,
\end{enumerate}
we are guaranteed that the solution $\hat{\Gama}_2$ for the Lagrangian formulation of the BP problem with parameter $\beta_2 = 4 \epsilon_1$ satisfies
\begin{enumerate}
\item The support of the solution $\hat{\Gama}_2$ is contained in that of $\Gama_2$;
\item $\| \Delt_2 \|_\infty = \| \Gama_2 - \hat{\Gama}_2 \|_{\infty} \leq 7.5 \ \epsilon_1$ \label{Eq:StabilitySoftITDistance};
\item In particular, every entry of $\Gama_2$ greater in absolute value than $7.5 \ \epsilon_1$ is guaranteed to be recovered; and
\item The solution $\hat{\Gama}_2$ is the unique minimizer of the Lagrangian BP problem (Equation~\eqref{Eq:LagrangianBP}).
\end{enumerate}
Using similar steps to those used above, the inequality that relies on the $\ell_\infty$ norm can be translated into another one that depends on the $\Ltinf$. This results in
\begin{equation}
\| \Delt_2 \|_{2,\infty}^\pp \leq \sqrt{ \| \Gama_2 \|_{0,\infty}^\pp } \ 7.5 \ \epsilon_1 = \epsilon_2.
\end{equation}
We conclude that, similar to the first one, the second stage of the layered BP is stable and the following must hold
\begin{enumerate}
\item The support of the solution $\hat{\Gama}_2$ is contained in that of $\Gama_2$;
\item $\| \Delt_2 \|_{2,\infty}^\pp \leq \ \epsilon_2$;
\item In particular, every entry of $\Gama_2$ greater in absolute value than $\frac{\epsilon_2}{\sqrt{ \| \Gama_{2} \|_{0,\infty}^\pp }}$ is guaranteed to be recovered; and
\item The solution $\hat{\Gama}_2$ is the unique minimizer of the Lagrangian BP problem (Equation~\eqref{Eq:LagrangianBP}).
\end{enumerate}
Using the same set of steps, we obtain similarly the stability of the remaining layers.
\end{proof}

\vskip 0.2in
\bibliography{MyBib}

\end{document}